\documentclass[10pt,twocolumn,letterpaper]{article}

\usepackage{iccv}
\usepackage{times}
\usepackage{epsfig}
\usepackage{graphicx}
\usepackage{amsmath}
\usepackage{amssymb}

\usepackage{subcaption}
\usepackage{tabularx}
\usepackage[usenames,dvipsnames]{xcolor}
\usepackage{booktabs}  %
\usepackage{fancyvrb}
\usepackage{fvextra}
\makeatletter
\@namedef{ver@everyshi.sty}{}
\makeatother
\usepackage{tikz}
\usetikzlibrary{shapes}

\usepackage[numbers,sort]{natbib}

\usepackage{xspace}
\def\eg{\emph{e.g.}\@\xspace}
\def\ie{\emph{i.e.}\@\xspace}
\def\etal{\emph{et al.}\@\xspace}
\def\cf{\emph{c.f.}\@\xspace}
\def\etc{\emph{etc}}

\renewcommand{\paragraph}[1]{\smallskip\noindent{\bf{#1}}}

\def\APall{mAP\textsubscript{all}\@\xspace}
\def\APrare{mAP\textsubscript{rare}\@\xspace}
\def\APcomm{mAP\textsubscript{comm}\@\xspace}
\def\APfreq{mAP\textsubscript{freq}\@\xspace}
\def\LVISminusrare{LVIS\textsubscript{-R}\@\xspace}

\def\zerorec{0Ways\@\xspace}
\def\onerec{1Ways\@\xspace}
\def\tworec{2Ways\@\xspace}
\def\threerec{3Ways\@\xspace}

\newcommand{\isArXiv}[2]{#1}
\usepackage[pagebackref=true,breaklinks=true,letterpaper=true,colorlinks,bookmarks=false]{hyperref}

\isArXiv{
\iccvfinalcopy
}{}

\def\iccvPaperID{1802}  %
\def\httilde{\mbox{\tt\raisebox{-.5ex}{\symbol{126}}}}

\isArXiv{
}{
\ificcvfinal\pagestyle{empty}\fi
}

\begin{document}

\def\fulltitle{Three ways to improve feature alignment for open vocabulary detection}
\title{\fulltitle}
\isArXiv{
\hypersetup{pdfauthor={Relja Arandjelović},pdftitle={\fulltitle}}
}
{
\hypersetup{pdfauthor={\iccvPaperID},pdftitle={\iccvPaperID: \fulltitle}}
}

\author{%
Relja Arandjelovi\'c\textsuperscript{1,*}
\quad
Alex Andonian\textsuperscript{2,$^*$,$^\circ$}
\\
Arthur Mensch\textsuperscript{1}
\quad
Olivier J.\ H\'enaff\textsuperscript{1}
\quad
Jean-Baptiste Alayrac\textsuperscript{1}
\quad
Andrew Zisserman\textsuperscript{1,3}
\\[0.5em]
$^1$DeepMind \quad {$^2$MIT} \quad {$^3$VGG, Dept.\  of Engineering Science, University of Oxford}
}

\maketitle
\isArXiv{
}{
\ificcvfinal\thispagestyle{empty}\fi
}

\newcommand{\meanstd}[2]{#1 {\tiny $\pm$ #2}}
\newcommand{\fail}[1]{\textcolor{BrickRed}{#1}}
\newcommand{\best}[1]{\textcolor{NavyBlue}{\bf{#1}}}
\newcommand{\second}[1]{\textcolor{RoyalPurple}{\underline{#1}}}

\newcommand{\figTeaser}{
\begin{figure}[t]
\vspace{-0.2cm}
\centering
\includegraphics[width=\linewidth]{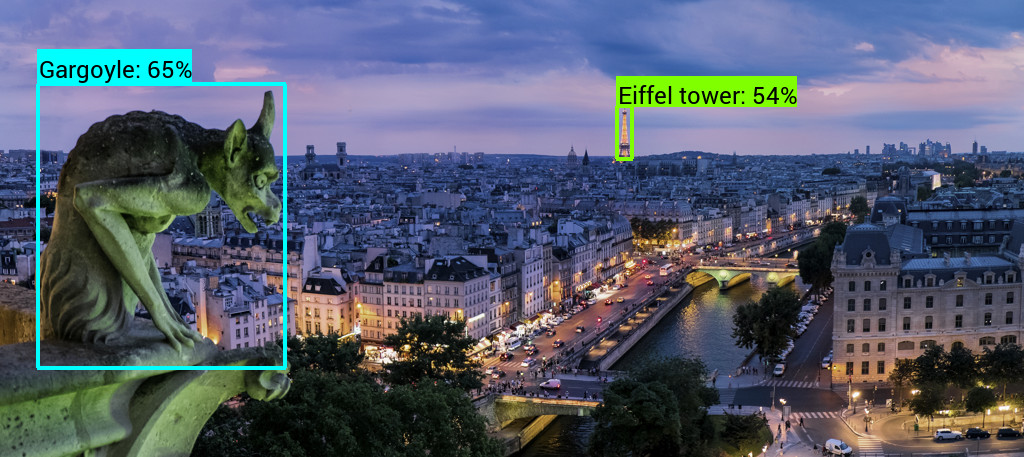}
\caption{{\bf Zero-shot open vocabulary detection.}
The detector is able to answer the queries ``Gargoyle'' and ``Eiffel tower''
despite never seeing human-annotated bounding boxes for them.
\vspace{-0.4cm}
}
\label{fig:teaser}
\end{figure}
}

\newcommand{\tabFreeze}{
\begin{table}[t]
\hspace*{-0.25cm}%
\centering
\begin{tabular}{lr@{~~}rcr} \toprule
LM train or freeze & \multicolumn{1}{c}{\small \APall} & \multicolumn{1}{c}{\small \APrare} & \multicolumn{1}{c}{\small speed} & \multicolumn{1}{c}{\small mem.} \\
\midrule
Train w/ lr-ratio 1 & \fail{\meanstd{12.0}{15.9}} & \fail{\meanstd{5.2}{7.30}} & 1.3 & 14.1G \\  %
Train w/ lr-ratio 0.01 & \best{\meanstd{33.2}{0.30}} & \meanstd{16.5}{0.95} & 1.3 & 14.1G \\  %
Freeze		& \fail{\meanstd{24.4}{13.2}} & \fail{\meanstd{13.0}{8.51}} & 1.9 & 10.4G \\  %
Freeze + Dropout & \meanstd{31.8}{0.20} & \best{\meanstd{18.7}{1.39}} & 1.7 & 10.4G \\  %
8 Templates, infer.\ 1 & \meanstd{31.3}{0.31} & \meanstd{16.4}{1.86} & \best{2.6} & \best{9.4G} \\  %
8 Templates, infer.\ 8 & \meanstd{31.5}{0.10} & \meanstd{17.1}{1.28} & \best{2.6} & \best{9.4G} \\  %
80 Templates, infer.\ 1 & \meanstd{31.6}{0.25} & \meanstd{17.4}{0.38} & \best{2.6} & \best{9.6G} \\  %
80 Templates, infer.\ 8 & \meanstd{31.9}{0.06}	& \meanstd{18.1}{0.66} & \best{2.6} & \best{9.6G} \\  %
1 Variant & \fail{\meanstd{16.3}{13.0}}	& \fail{\meanstd{6.9}{7.89}} & \best{2.6} & \best{9.4G}	\\  %
64 Variants & \meanstd{32.1}{0.31} & \best{\meanstd{18.9}{1.13}} & \best{2.6} & \best{9.5G} \\  %
\bottomrule
\end{tabular}
\caption{{\bf To train or to freeze the language model (\LVISminusrare benchmark).}
Speed is measured as the number of gradient steps per second,
while `mem.' denotes the peak accelerator memory usage.
Methods where at least 1 out of the 3 training runs has failed are
in \fail{red}.
Our \emph{Freeze + Dropout} and \emph{64 Variants} approaches
perform best on the unseen classes, while speeding up training
and requiring less memory.  %
}
\label{tab:freeze}
\vspace{-0.2cm}
\end{table}
}

\newcommand{\tabInit}{
\def\y{\checkmark}
\def\n{~}
\begin{table}[t]
\centering
\begin{tabular}{cc@{~~}c@{~~}cc@{~~}c@{~~}cc} \toprule
\multicolumn{2}{c}{Architecture} & ~ & \multicolumn{2}{c}{APA} & ~ & \multicolumn{2}{c}{\LVISminusrare performance} \\
\cmidrule{1-2} \cmidrule{4-5} \cmidrule{7-8}
Backb. & Head && FPN & Head & & \APall & \APrare \\
\midrule
NF-F0 & FCOS && \n & \n && \meanstd{30.4}{0.21} & \meanstd{16.1}{1.46} \\  %
NF-F0 & FCOS && \y & \y && \best{\meanstd{32.7}{0.15}} & \best{\meanstd{19.8}{0.34}} \\  %
\midrule
NF-F0 & T-Head && \n & \n && \meanstd{32.1}{0.31} & \meanstd{18.9}{1.13} \\  %
NF-F0 & T-Head && \y & \n && \meanstd{32.4}{0.44} & \meanstd{18.3}{1.58} \\  %
NF-F0 & T-Head && \n & \y && \meanstd{33.3}{0.15} & \meanstd{19.6}{0.49} \\  %
NF-F0 & T-Head && \y & \y && \best{\meanstd{33.8}{0.15}} & \best{\meanstd{20.9}{0.34}} \\  %
\midrule
NF-F6 & T-Head && \n & \n && \meanstd{41.6}{0.17} & \meanstd{21.1}{0.40} \\  %
NF-F6 & T-Head && \y & \y && \best{\meanstd{43.5}{0.12}} & \best{\meanstd{27.6}{0.80}} \\  %
\bottomrule
\end{tabular}
\caption{{\bf Alignment preserving architecture (APA).}
All networks were trained with the \emph{64 Variants} approach (Section~\ref{sec:aug}).
The added shortcuts and trainable gating layers consistently improve
the detection performance for both the backbone and detection head architectures.
}
\label{tab:gating}
\vspace{-0.2cm}
\end{table}
}

\newcommand{\tabSelfTrain}{
\begin{table}[t]
\hspace*{-0.25cm}%
\centering
\begin{tabular}{llrr} \toprule
Self-training method & Backb. & \multicolumn{1}{c}{\APall} & \multicolumn{1}{c}{\APrare} \\
\midrule
\tworec (no self-training) & NF-F0 & \meanstd{33.8}{0.15} & \meanstd{20.9}{0.34} \\  %
Image bbox w/o batch-negs & NF-F0 & \meanstd{33.9}{0.30} & \meanstd{20.9}{1.06}  \\ %
Image bbox & NF-F0 & \meanstd{35.1}{0.26} & \meanstd{24.2}{1.37} \\  %
Detic~\cite{zhou2022detecting} \scriptsize{open-voc.} $^\dagger$ & NF-F0 & \meanstd{34.8}{0.32} & \meanstd{23.4}{1.49} \\  %
\threerec w/o batch-negs & NF-F0 & \meanstd{34.2}{0.15} & \meanstd{20.7}{0.35}  \\ %
\threerec & NF-F0 & \best{\meanstd{35.7}{0.20}} & \best{\meanstd{25.6}{1.12}} \\  %
\midrule
\tworec (no self-training) & NF-F6 & \meanstd{43.5}{0.12} & \meanstd{27.6}{0.80} \\  %
\threerec & NF-F6 & \best{\meanstd{44.6}{0.31}} & \best{\meanstd{30.1}{1.83}} \\  %
\bottomrule
\end{tabular}
\caption{{\bf Self-training (\LVISminusrare benchmark).}
CC12M is pseudo-labelled with the \emph{\tworec} detector (Sections~\ref{sec:aug} and~\ref{sec:init}).
Detic$^\dagger$ is our reimplementation of Detic~\cite{zhou2022detecting},
Image bbox uses the entire image as the pseudo-detection while
\threerec uses the \tworec's best prediction;
Section~\ref{sec:selft} explains all methods.
Self-training helps, and batch-negatives are important.
}
\label{tab:selftrain}
\vspace{-0.2cm}
\end{table}
}

\newcommand{\tabSota}{
\def\y{\checkmark}
\def\n{~}
\begin{table*}[t]
\vspace{-0.4cm}
\centering
\begin{tabularx}{\linewidth}{>{\hsize=1.2\hsize\linewidth=\hsize}X >{\hsize=0.8\hsize\linewidth=\hsize}X rcllll}
\toprule
Method & Backbone & \multicolumn{1}{c}{\#Params} & Self-training & \multicolumn{1}{c}{\APall} & \multicolumn{1}{c}{\APrare} & \multicolumn{1}{c}{\APcomm} & \multicolumn{1}{c}{\APfreq} \\
\midrule
Detic~\cite{zhou2022detecting} \scriptsize{open-voc.}$^{(m)}$ & R50 & 26M & \y & 30.4 & 17.4 \\
DetPro~\cite{du2022learning} & R50 & 26M & \n & 28.4 & 20.8 & 27.8 & 32.4 \\
RegionCLIP~\cite{zhong2022regionclip} & R50x4 & 87M & \y & 32.1 & 22.0 & 32.1 & 36.9 \\
OWL~\cite{minderer2022simple} & VIT-L/14 & 303M & \n & 34.7 & 25.6 \\
OWL~\cite{minderer2022simple} & VIT-H/14 & 627M & \n & 35.3 & 23.3 \\
F-VLM~\cite{kuo2023fvlm} & R50x4 & 87M & \n & 28.5 & 26.3 \\
F-VLM~\cite{kuo2023fvlm} & R50x64 & 420M & \n & 34.9 & \best{32.8} \\
\zerorec \scriptsize{[this work]} & NFNet-F0 &  71M & \n & \meanstd{16.3}{13.0}	& \multicolumn{1}{r}{\meanstd{6.9}{7.89}} & \meanstd{13.2}{11.3} & \meanstd{23.7}{11.7} \\  %
\onerec \scriptsize{[this work]} & NFNet-F0 &  71M & \n & \meanstd{32.1}{0.31} & \meanstd{18.9}{1.13} & \meanstd{29.5}{0.15} & \meanstd{40.9}{0.08} \\  %
\tworec \scriptsize{[this work]} & NFNet-F0 &  71M & \n & \meanstd{33.8}{0.15} & \meanstd{20.9}{0.34} & \meanstd{32.4}{0.20} & \meanstd{41.0}{0.05} \\  %
\threerec \scriptsize{[this work]} & NFNet-F0 &  71M & \y & \meanstd{35.7}{0.20} & \meanstd{25.6}{1.12} & \meanstd{34.2}{0.05} & \meanstd{41.8}{0.02} \\  %
\zerorec \scriptsize{[this work]} & NFNet-F6 & 440M & \n & \multicolumn{1}{r}{\meanstd{0.8}{0.19}} & \multicolumn{1}{r}{\meanstd{0.4}{0.12}} & \multicolumn{1}{r}{\meanstd{0.7}{0.16}} & \multicolumn{1}{r}{\meanstd{1.0}{0.24}} \\  %
\onerec \scriptsize{[this work]} & NFNet-F6 & 440M & \n & \meanstd{41.6}{0.17} & \meanstd{21.1}{0.40} & \meanstd{42.9}{0.19} & \meanstd{49.2}{0.09} \\  %
\tworec \scriptsize{[this work]} & NFNet-F6 & 440M & \n & \meanstd{43.5}{0.12} & \meanstd{27.6}{0.80} & \meanstd{44.9}{0.10} & \meanstd{48.8}{0.01} \\  %
\threerec \scriptsize{[this work]} & NFNet-F6 & 440M & \y & \best{\meanstd{44.6}{0.31}} & \meanstd{30.1}{1.83} & \best{\meanstd{46.0}{0.17}} & \best{\meanstd{49.3}{0.08}} \\  %
\bottomrule
\end{tabularx}
\caption{{\bf State-of-the-art for zero-shot open vocabulary detection on \LVISminusrare.}
$^{(m)}$ denotes that Detic~\cite{zhou2022detecting} only reports the mask mAPs,
box mAP should not be much higher.
\emph{\zerorec}, \emph{\onerec}, \emph{\tworec} and \emph{\threerec}
refer to
the baseline detector with the frozen LM and no text augmentation,
and cumulative application of our three methods
from Sections~\ref{sec:aug},~\ref{sec:init} and~\ref{sec:selft}, respectively.
\emph{\threerec} performs well, yielding the best \APall by a large margin
and achieving a favourable \APrare.
}
\vspace{-0.2cm}
\label{tab:sota}
\end{table*}
}

\newcommand{\figSetup}{
\begin{figure*}[t]
\definecolor{lang}{HTML}{134f5c}
\definecolor{vis}{HTML}{1155cc}
\definecolor{det}{HTML}{b45f06}
\vspace{-0.6cm}
\includegraphics[width=\linewidth]{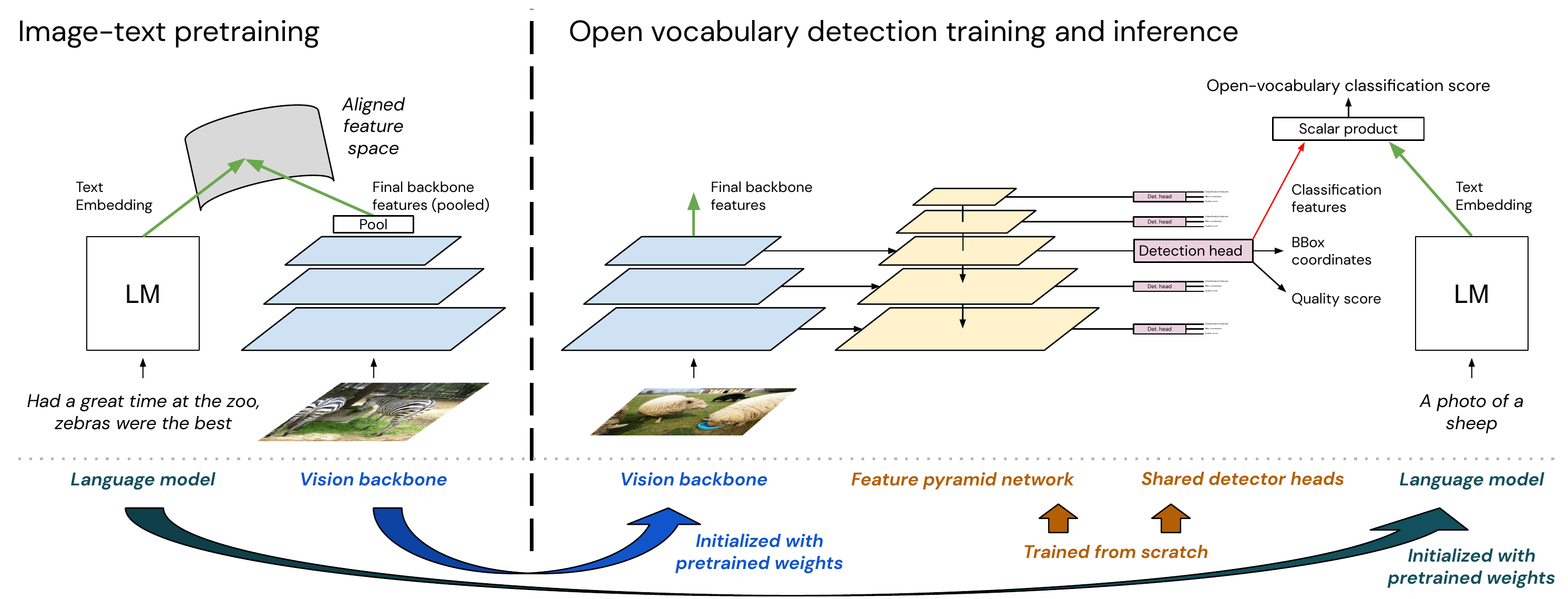}
\caption{{\bf A standard approach to open vocabulary detection and pretraining.}
A standard single-stage detector adapted to open vocabulary detection,
as explained in Section~\ref{sec:setup},
makes use of a \textcolor{lang}{language model}, \textcolor{vis}{vision backbone}, \textcolor{det}{feature pyramid network (FPN)},
and \textcolor{det}{detector heads}.
The \textcolor{vis}{vision backbone} and \textcolor{lang}{language model} are typically pretrained in
a contrastive manner, while the \textcolor{det}{FPN} and the \textcolor{det}{detector heads} are initialized
from scratch.
}
\label{fig:setup}
\vspace{-0.2cm}
\end{figure*}
}

\newcommand{\figGating}{
\begin{figure*}[t]
\definecolor{arrow}{HTML}{6aa84f}
\definecolor{gate}{HTML}{ff9900}
\definecolor{back}{HTML}{cfe2f3}
\DeclareRobustCommand{\orangehexagon}{$\mathord{\raisebox{0.6pt}{\tikz{\node[draw,scale=.65,regular polygon, regular polygon sides=6,fill=gate](){};}}}$\@\xspace}
\includegraphics[width=\linewidth]{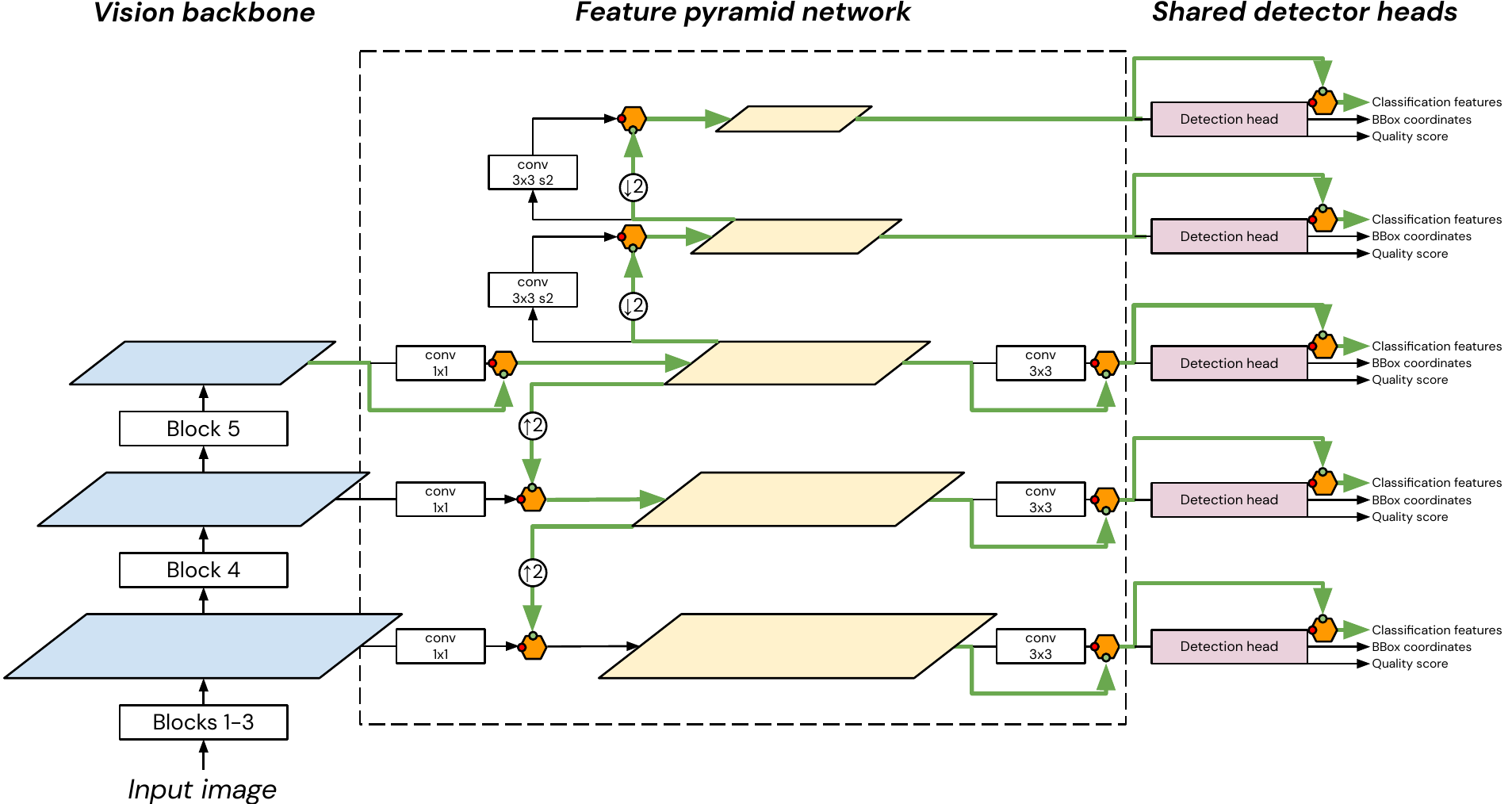}
\caption{{\bf Alignment preserving architecture (APA).}
The standard single-stage object detector architecture
[Backbone → Feature pyramid network (FPN) → Detector heads]
is augmented with shortcuts and trainable gating layers
(\orangehexagon),
which at init propagate
the \textcolor{arrow}{green} input and block the \textcolor{red}{red} input.
The output is computed as
$\textcolor{arrow}{x} (1-\tan\alpha) + \textcolor{red}{y} \tan{\alpha}$,
and $\alpha=0$ at init.
The \textcolor{arrow}{green arrows} show the propagation of the last
backbone features at init all the way to the final
detector head classification features.
\textcolor{CornflowerBlue}{Light blue} and \textcolor{Apricot}{light yellow} parallelograms
represent the backbone and FPN feature maps, respectively;
circles with ↑2 and ↓2 are non-trainable up- and down-sampling,
squares are trainable modules (\eg convolutions),
and \orangehexagon are the trainable gates;
convolution blocks show the kernel size
and potential striding (`s2': stride 2).
The standard architecture (\ie without the shortcuts and gates) is shown
in the supplementary material.
}
\label{fig:gating}
\vspace{-0.2cm}
\end{figure*}
}

\newcommand{\tabTransfer}{
\begin{table}[t]
\centering
\begin{tabular}{llrcc} \toprule
Method & Backbone & \#Params & COCO & O365 \\
\midrule
ViLD~\cite{gu2022openvocabulary} & R50 & 26M & 36.6 & 11.8 \\
DetPro~\cite{du2022learning} & R50 & 26M & 34.9 & 12.1 \\
F-VLM~\cite{kuo2023fvlm} & R50x4 & 87M & 36.0 & 14.2 \\
F-VLM~\cite{kuo2023fvlm} & R50x64 & 420M & 39.8 & 17.7 \\
\tworec \scriptsize{[this work]} & NF-F0 & 71M & 40.6 & 14.6 \\
\threerec \scriptsize{[this work]} & NF-F0 & 71M & 41.5 & 16.4 \\
\tworec \scriptsize{[this work]} & NF-F6 & 440M & 46.5 & 20.3 \\
\threerec \scriptsize{[this work]} & NF-F6 & 440M & \best{46.9} & \best{22.8} \\
\bottomrule
\end{tabular}
\caption{{\bf Transfer.}
The \LVISminusrare trained networks are evaluated
on COCO~\cite{lin2014coco} and Objects365-v1~\cite{shao2019objects365} without any additional training.
}
\label{tab:transfer}
\vspace{-0.3cm}
\end{table}
}

\begin{abstract}
The core problem in zero-shot open vocabulary detection is how to align
visual and text features, so that the detector performs well on unseen
classes. Previous approaches train the feature pyramid and detection head
from scratch, which breaks the vision-text feature alignment established
during pretraining, and struggles to prevent the language model from
forgetting unseen classes.

We propose three methods to alleviate these issues. Firstly, a simple
scheme is used to augment the text embeddings which prevents overfitting
to a small number of classes seen during training, while simultaneously
saving memory and computation. Secondly, the feature pyramid network and
the detection head are modified to include trainable gated shortcuts,
which encourages vision-text feature alignment and guarantees it at
the start of detection
training. Finally, a self-training approach is used to leverage a larger
corpus of image-text pairs thus improving detection performance on classes
with no human annotated bounding boxes.

Our three methods are evaluated on the zero-shot version of the LVIS benchmark,
each of them showing clear and significant benefits. Our final network
achieves the new state-of-the-art on the mAP-all metric and demonstrates
competitive performance for mAP-rare,
as well as superior transfer to COCO and Objects365.
\vspace{-0.4cm}
\end{abstract}

\isArXiv{
{\let\thefootnote\relax\footnotetext{$^*$Equal contribution. $^\circ$Work done during internship at DeepMind.}}
}{}

\section{Introduction}
\label{sec:intro}

Traditional closed vocabulary detection is limited to a fixed set of
predetermined classes, and does not satisfactorily address user needs
-- imagine Google where you are only able to search for
a predefined list of terms.
Adding support for more terms requires large and costly
annotation efforts, which is simply not scalable.
Our objective in this paper is zero-shot open vocabulary detection, where the task is to
detect any object the user queries for, in a form of a textual query
(\eg ``Gargoyle''; Figure~\ref{fig:teaser}),
even if it has not been seen during training.

The common approach to building an open vocabulary detector is to
borrow heavily from the design of standard closed vocabulary detectors
(\ie detectors capable of detecting only a fixed set of predetermined classes),
and simply modify the bounding box classification procedure.
Instead of producing the logits for the fixed set of classes
via a fully connected layer, the score for the textual query is
obtained via a scalar product between its embedding,
produced by a language model,
and the image region embedding, produced by the detector head.
The zero-shot capability strongly relies on a good alignment between
visual and textual representations --
the only way queries not seen during training can be detected successfully
is if the vision-text alignment holds even beyond the seen classes.
In this work, we explicitly consider feature alignment
and devise three ways of improving it:

(i) Many works~\cite{zareian2020openvocabulary,zhou2022detecting,kuo2023fvlm}
choose to freeze the pretrained language model, while others,
observe this yielding bad performance~\cite{minderer2022simple,li2022glip,zhang2022glipv2}
and choose to train it, but with a small learning rate
to prevent ``catastrophic forgetting''. We also find that a frozen language model alone yields poor performance (Section~\ref{sec:aug:res}), but propose instead
to use the frozen language model together with a simple and
efficient data augmentation approach, which provides superior results to
both alternatives while speeding up training and decreasing accelerator
memory consumption.

\figTeaser
\figSetup

(ii) A typical detector pretrains the vision backbone
and language model on image-text datasets to obtain aligned image and text embeddings~\cite{zhou2022detecting,feng2022promptdet,gu2022openvocabulary,minderer2022simple},
but also inserts many modules
(feature pyramid network~\cite{lin2017feature},
detection heads~\cite{lin2017feature,tian2019fcos,feng2021tood})
that are trained from scratch.
The added modules break the vision-text alignment established during pretraining,
and we propose to side-step this issue by modifying their architecture.
Explicitly, we add shortcuts and trainable gating layers which ensure the features
are aligned at the start of detector training, and promote alignment
throughout the training.

(iii) Feature alignment that can be achieved from relatively scarce
detection training data is sparse and limited. The alignment can be improved by
making use of readily available large-scale image-text data
through a self-training approach~\cite{xie2020selftraining,lee2013pseudolabel,sohn2020simple,zoph2020rethinking}.
We examine self-training
via pseudo-labelling in detail and observe it is crucial
to use batch-negatives.

Our final approach based on all three improvements achieves the best
\APall on the challenging \LVISminusrare benchmark, beating the next method
by more than 9\% points, while achieving very competitive zero-shot
results and superior transfer to COCO and Objects365.

\subsection{Related work}

\paragraph{Zero-shot open vocabulary detection.}
Zero-shot (ZS) in the context of object detection
refers to never seeing even a single
annotated bounding box of the class of interest during training~\cite{minderer2022simple};
note that this definition allows for the existence of the object in the
training set images as long as no annotations are associated with it,
and it permits weak supervision, \eg an image-text dataset
where the object is mentioned in the text can be used as long as no bounding boxes are provided.
There is a large overlap in the ZS and open vocabulary (OV) approaches,
so, confusingly, the terms are often used interchangeably, which we avoid here.

Bansal \etal~\cite{bansal2018zeroshot} introduce ZS+OV detection
where the classification layer of a closed vocabulary detector
is replaced with the text embeddings of the class names,
an approach taken by many subsequent works~\cite{zhou2022detecting,du2022learning,minderer2022simple,zareian2020openvocabulary,gu2022openvocabulary,zhou2022detecting,kuo2023fvlm,feng2022promptdet},
including this one.
Some works~\cite{zareian2020openvocabulary,gu2022openvocabulary,kuo2023fvlm}
take the OV classification closer to the
backbone features by directly extracting them from object proposals
with ROI-Align~\cite{he2017mask}, and optionally distill a strong
OV classifier into the detector~\cite{gu2022openvocabulary}.
To improve ZS performance, Detic~\cite{zhou2022detecting} and PromptDet~\cite{feng2022promptdet} forego the OV aspect --
knowing the names of the classes of interest
(\ie in evaluation: test classes)
already during training
enables them to obtain high-quality weak labels,
and thus improve the detection performance for those classes.

\paragraph{Self-training} is often used in the weakly- and semi-supervised
settings to improve the low-shot performance of a detector~\cite{redmon2017yolo9000,sohn2020simple,zoph2020rethinking,ramanathan2020dlwl},
by first training a detector, followed by using it to pseudo-label additional
images, which are in turn used to train a better detector.
This has been adapted by Detic~\cite{zhou2022detecting}
to the ZS scenario, who argue for using region proposals
rather than the detector outputs to perform the pseudo-labelling.
Motivated by self-supervision and contrastive learning~\cite{oord2018representation,chen2020simple,arandjelovic2017look,he2020momentum,radford2021learning,alayrac2022flamingo},
we show that using batch-negatives is crucial for obtaining good performance
in self-training as well.

\paragraph{Pretraining-preserving init.}
The seminal ResNet paper~\cite{He16} showed the importance of shortcuts
for signal propagation during training, while
SkipInit~\cite{de2020batch} introduced a learnt gating that further encourages
identity functions. We take most inspiration from
the trainable gating of Flamingo~\cite{alayrac2022flamingo}
where the vision-language model is initialized such that the visual branch
is ignored, thus preserving the language model pretraining.
FIBER~\cite{dou2022coarsetofine} uses the Flamingo-style gating
to initialize a joint vision-text encoder with pretrained dual encoders.
We instead aim to preserve alignment between visual and language
features obtained during pretraining but broken due to the injected
detection-specific modules.

\section{Baseline detector and experimental setup}
\label{sec:setup}

In this section, we describe the baseline
open vocabulary detector (Figure~\ref{fig:setup}),
that we build and improve upon in Section~\ref{sec:rec}.
We also specify the main benchmark with some implementation details,
while the full details are available in \isArXiv{Appendix~\ref{sec:app:imp}.}{the supplementary material.}

\paragraph{Open vocabulary detector.}
We follow the design of the single-stage FCOS detector~\cite{tian2019fcos},
illustrated in Figure~\ref{fig:setup}.
It starts by processing the image with a vision backbone,
features from different blocks are then passed to the feature pyramid network
(FPN~\cite{lin2017feature}),
followed by the application of detection heads (parameters shared across levels);
we use the T-Head~\cite{feng2021tood} but also experiment with the classic
FCOS head~\cite{tian2019fcos}.
Each head produces dense detections associated with three quantities:
bounding box coordinates, quality, and classification features.
In line with other ZS/OV approaches~\cite{bansal2018zeroshot,zhou2022detecting,zareian2020openvocabulary,gu2022openvocabulary},
the classification features are dotted with the embedding of the query text,
obtained via a language model (LM),
producing the classification logits for the given query.
The final scores for all dense detections are computed by multiplying the classification probabilities with the quality scores.
Non-maximum suppression~\cite{felzenszwalb10object} is then applied to produce the final detections.
Training follows the standard FCOS method and its improvements,
\ie
the dense predictions are assigned to a ground truth box or deemed
as a negative through ATSS~\cite{zhang2020bridging},
and the classification, bounding box prediction, and
quality prediction branches
use the focal~\cite{lin2017focal}, gIoU~\cite{rezatofighi2019generalized}
and IoU prediction~\cite{wu2020iou} losses, respectively.

Free form textual queries are naturally supported,
while it is still possible
to detect a desired object class as
the query text for that class (hereafter also referred to as the ``class embedding'')
can be produced by populating
the default template (``A photo of a \emph{\{object\}}'')
with the class name.

\paragraph{Zero-shot benchmark.}
We use the LVIS v1.0~\cite{gupta2019lvis} object detection benchmark
adapted for zero-shot evaluation; we call this setup \LVISminusrare.
Following standard practice~\cite{gu2022openvocabulary,zhou2022detecting,kuo2023fvlm},
the \emph{rare} class annotations are removed from the training set, keeping
only the \emph{frequent} and \emph{common} annotations (often called LVIS-base).
Evaluation is then performed on all classes, reporting the
box mAP for all classes (\APall) and the mAP on rare classes (\APrare),
with the emphasis on \APrare as this measures the zero-shot performance,
\emph{rare} classes playing the role of the \emph{unseen} classes.
As is best practice~\cite{lvisbestpractices}, we run all experiments with three
different random seeds and report the mean and standard deviation.

\paragraph{Implementation details.}
The vision backbone and the language model are pretrained contrastively
on the ALIGN~\cite{jia2021scaling} and LTIP datasets~\cite{alayrac2022flamingo}
as in~\cite{alayrac2022flamingo},
while the FPN and the detector head are initialized
from scratch.
We follow a standard training procedure for \LVISminusrare
and tune the hyper-parameters to maximize the baseline performance;
full details are listed in \isArXiv{Appendix~\ref{sec:app:imp}.}{the supplementary material.}
With the NFNet-F0~\cite{brock2021highperformance} backbone
we achieve an \APall of \meanstd{32.1}{0.31}, and
\APrare of \meanstd{18.9}{1.13}.
This is a strong baseline,
as for example the baseline used in a recent work~\cite{zhou2022detecting}
achieves \meanstd{30.0}{0.4} and \meanstd{16.3}{0.7}, respectively.

\section{Three paths to alignment}
\label{sec:rec}

In this section we describe three complementary methods for improving
vision-text alignment,
starting from efficient text augmentation which alleviates overfitting
and facilitates large scale training (Section~\ref{sec:aug}),
followed by an architectural modification that preserves and promotes the alignment
(Section~\ref{sec:init}),
and ending with an approach for self-training which further improves
the detection performance on unseen classes (Section~\ref{sec:selft}).

\subsection{Efficient text augmentation}
\label{sec:aug}

When training a zero-shot detector, a difficult choice has to be made
whether to train or to freeze the language model (LM).
Many works, such as OVD~\cite{zareian2020openvocabulary}, Detic~\cite{zhou2022detecting}
and F-VLM~\cite{kuo2023fvlm},
follow the natural intuition to freeze it --
the language model learnt a comprehensive textual representations during
pretraining, and fine-tuning for detection on a small number of classes
could make it forget about the unseen classes~\cite{minderer2022simple}.
However, freezing it also comes with downsides --
the vision model is ``forced'' into the language-model ``mould''
making it less able to adapt to the task change from
pretraining which only involved global image understanding.
Multiple works, such as OWL~\cite{minderer2022simple}, GLIP~\cite{li2022glip}, GLIPv2~\cite{zhang2022glipv2},
do train the language model as well, but typically use a smaller learning
rate in order to prevent ``catastrophic forgetting'', \eg OWL~\cite{minderer2022simple}
sets it to $1/100$ of the vision model learning rate
and notes poor performance when the language model is frozen.

In fact, experimentally we find that the main issue behind the poor
detection performance of a system with a frozen language model is
overfitting of the visual representations to the small fixed set of
textual embeddings corresponding to the training classes.
Augmenting of the class embeddings during training can be used as an
effective way of alleviating these issues, and we consider two alternatives:

\noindent (i) \emph{Freeze + Dropout}: despite freezing the LM, enable the
dropout (commonly present in Transformer-based LMs during training).

\noindent (ii) \emph{Variants}: precompute 64 variants of the class embeddings
by using (i) and randomly sample a variant for each training sample.

Freezing the LM makes the training faster and simultaneously saves memory
due to not having to perform backpropagation or keep the optimizer state
(\eg for popular stateful optimizers such as SGD with momentum or Adam~\cite{Kingma15}).
The \emph{Variants} approach makes it possible to completely remove the LM
during training as precomputed embeddings can be used,
thus making training even faster and providing further memory savings.
This can be essential as detection training requires high-resolution images
which for some large vision models makes it hard to fit even
a batch size of 1 into the accelerator memory --
it is exactly the case for our self-training NFNet-F6 experiments
(Section~\ref{sec:selft})
which are not possible without the \emph{Variants} method.

\vspace{-0.15cm}
\subsubsection{Results}
\vspace{-0.1cm}
\label{sec:aug:res}

\tabFreeze

All experiments are performed on the \LVISminusrare benchmark (\cf Section~\ref{sec:setup}).
The effect of different approaches to training or freezing the language model
are shown in Table~\ref{tab:freeze}.

\paragraph{Training}
the LM with the same learning rate as the vision model is unstable
and results in poor performance.
Using the OWL~\cite{minderer2022simple} strategy of training the LM with a very small
learning rate yields good performance on all classes.
However,
when compared with our augmentation approaches,
it becomes clear that it is underperforming on unseen classes,
as even this small learning rate
causes some forgetting, albeit arguably not ``catastrophic''.

\paragraph{Freezing}
the LM underperforms and is unstable, testifying to overfitting;
this is equivalent to the \emph{1 Variant} scenario which performs
equally badly.
However, simply using dropout while keeping the network frozen performs
very well -- achieving the best mAP on the unseen classes.
Furthermore, the \emph{64 Variants} approach, where the variants of
the class embeddings are precomputed and sampled during training, performs
equally well while
enabling us to remove the LM inference from training.
This in turn achieves
a 9\% reduction in memory use and a speedup of 53\% vs \emph{Freeze + Dropout},
and 33\% memory savings and a 2$\times$ speedup vs the LM-training approaches.

\figGating

\paragraph{Templates.}
An alternative to the \emph{Variants} approach is to use
many different text templates
(\eg ``A close-up photo of the \emph{\{object\}}'' or ``A low resolution photo of the \emph{\{object\}}'')
to compute the
class embeddings and randomly sample them for each training sample~\cite{minderer2022simple}.
The 8 templates are formed by combining the ``7 best'' CLIP templates~\cite{radford2021learning}
and the default one (``A photo of a \emph{\{object\}}''), while the 80 templates are
the CLIP 80 templates (the default is already included).
The use of multiple templates during training has a similar effect to
\emph{64 Variants} in that it trains stably and outperforms
the LM-training approaches.
However, for good performance it requires inference with multiple
templates~\cite{minderer2022simple}
(\ie class probabilities are averaged across the different templates)
which increases complexity and
inference memory requirements, while still being beaten by our simple
\emph{Variants} approach.
We hypothesise this is because the templates have been designed for
ImageNet classification and
contain obscure concepts such as ``An origami \emph{\{object\}}''.
It is not easy to design many good templates, so our approach to
simply compute \emph{64 Variants} of the class embedding by using the
natural default template and enabling dropout is more effective.

\subsection{Alignment preserving architecture}
\label{sec:init}

As outlined in Section~\ref{sec:intro} and shown in Figure~\ref{fig:setup},
a typical setup that we also follow is to:
(1) pretrain a vision-language model,
(2) construct the open-vocabulary detector by re-using the vision and
text backbones and adding detection-specific layers
(feature pyramid network (FPN) and detector heads),
(3) initialize the backbones from (1) while initializing the rest
(FPN, heads) from scratch, and
(4) train all or subsets of parameters \eg freezing the LM backbone.

The disconnect between steps (1) and (3) stands out --
the vision and language backbones were trained together to produce aligned
representations of their respective modalities,
and we sever that alignment by introducing many layers in-between that are
trained from scratch.
The detector training then spends a long time seeking to realign
the features, and it is very likely that during this initial chaos some
of the pretrained alignment is forever lost.
Here we introduce small architectural changes
to the detector-specific layers
which serve to maintain the alignment of vision-text features at the start,
and promote it throughout the detector training.

The architectural modifications are shown in Figure~\ref{fig:gating} and consist
of strategically adding shortcut connections and
trainable gating layers~\cite{alayrac2022flamingo}.
A trainable gating layer,
with inputs $x$ and $y$ and a trainable scalar parameter $\alpha$,
produces the output $o = x (1-\tan\alpha) + y \tan{\alpha}$,
where $\alpha$ is initialized to 0 meaning $o = x$ at the start of training.
The shortcuts and gates are added such that at
the start of training, the features from the end of the vision backbone
are ``forwarded'' through the FPN and detector heads
all the way to the final classification features.
In other words, at the start of training, the detector head classification
features at all levels of the pyramid are equal to the backbone features.
Recall that the vision and text backbones have been pretrained for alignment.
This means that due to the specific gated-shortcut architecture and
initialization, the detection head classification features
(now equal to the backbone features)
are already aligned with the text embeddings at the beginning of
detector training.
Thus, the training is improved as it starts from a good initial guess
for the object classification and only needs to learn to improve
the classification and bounding box prediction,
rather than spend effort in rediscovering the vision-text alignment.

\paragraph{Aligned architecture design.}
Here we explain in more detail the recipe for converting an architecture
to its gated-shortcut version.
As explained above, the overall aim is to forward the final backbone features
(as they are pretrained to be aligned with the text embeddings)
to the end of the detection head.
So one only needs to follow the ``flow'' of the final backbone features
and apply the following operations:
(i) if they are mixed with another signal, add a gate that zeroes-out
the second signal at the start of training,
(ii) if an alignment preserving operation is performed (\eg upsampling)
do nothing,
(ii) if an alignment damaging transformation is performed (\eg a convolution),
make a shortcut connection and
add a gate such that the output equals the shortcut at the start of training.

These principles and the resulting architecture are illustrated in
Figure~\ref{fig:gating},
where the FPN is augmented with the shortcuts and gates,
while a single shortcut+gate combination is used around the entire detector head.
This makes it easy to apply the design to different detector heads
(\eg FCOS~\cite{tian2019fcos} vs T-Head~\cite{feng2021tood}) which contain
potentially more complex operations.

\vspace{-0.2cm}
\subsubsection{Results}
\label{sec:init:res}

\tabInit

Table~\ref{tab:gating} shows the results of our the alignment preserving
architectures (APA).

\paragraph{Alignment preserving vs vanilla architecture.}
Coupled with the NFNet-F0 vision backbone and the FCOS~\cite{tian2019fcos}
detector head, our design improves
\APall and \APrare by +2.3\% and +3.7\%, respectively.
Similarly, for the better performing T-Head~\cite{feng2021tood}
APA achieves +1.7\% and +2\%, respectively.
It is impressive that the improvements transfer to the larger
NFNet-F6 network which already exhibits an excellent \APall of 41.6\%,
which is further boosted by APA by +1.9\% to reach 43.5\%.
The largest improvement can be observed for \APrare where
the alignment preserving architecture tops the strong baseline by +6.5\%
and yields 27.6\%.

\paragraph{In the FPN, Head or everywhere?}
Table~\ref{tab:gating} shows it is more important to apply APA onto
the detection head than the FPN -- we speculate that this is because
the detection head is much deeper and therefore without APA
it takes longer to learn to re-learn the feature alignment
in the detector head than in the FPN.
However, applying APA onto both simultaneously clearly dominates,
confirming our intuition that maintaining alignment from the very
start of training is important.

\subsection{Self-training}
\label{sec:selft}

Text augmentation (Section~\ref{sec:aug}) and the
alignment preserving architecture (Section~\ref{sec:init})
bring significant gains in zero-shot performance due to the improved
feature alignment. However, it is still ambitious to ask for the detector
to extrapolate to completely unseen classes.
In this section, we investigate how to use self-training via pseudo-labelling
to further improve the feature alignment beyond the seen classes.

We propose a simple three-stage approach.
First, a good open vocabulary detector is trained using the previous
two improvements (Sections~\ref{sec:aug} and~\ref{sec:init}), called \emph{\tworec}.
The detector is then used to pseudo-label an additional dataset that
contains only images-text pairs scraped from the internet,
\ie it contains weak image-level information (the text), without any
human supervision nor finer-grained annotations such as classes,
bounding boxes or segmentations.
The detector uses the text embedding of the entire caption as the object query,
and we simply use the single highest scoring box per image if it passes a
confidence threshold of 0.25.
Finally, a new, stronger, open vocabulary detector (\emph{\threerec}) is trained
by combining the strongly supervised data (\LVISminusrare) and the
pseudo-labelled dataset, and treating the pseudo-labels as ground truth.

It is worth elaborating on the exact details of the final training stage.
Recall from Section~\ref{sec:setup} that for training with the true
ground truth annotations, we follow
the standard training procedure; \ie, certain detector head classification
features are assigned to be positives for particular classes
(in the open vocabulary case, its text embedding)
based on their pyramid level and location in the feature map~\cite{zhang2020bridging}.
The same features are negatives for other classes, and all remaining features
are negatives for all classes.
For example, if an image has a \emph{dog} in it and no \emph{cat}, we have:
(i) some features depending on scale and location are positives for \emph{dog},
(ii) features that are not positives for \emph{dog} are negatives for \emph{dog},
(iii) all features are negatives for \emph{cat}.
Training then proceeds with the standard per-class binary focal loss~\cite{lin2017focal}.

We propose an analogous mechanism when training with the pseudo-labels.
The single pseudo-bounding box per image is deemed to correspond to the entire caption,
and other captions in the batch are used as negatives.
Therefore, for the $i$-th image in the batch, we have:
(i) some features are deemed positive for the $i$-th caption
again following~\cite{zhang2020bridging},
(ii) features that are not positive are negatives for the $i$-th caption,
(iii) all features are negatives for the $j$-th caption where $i \neq j$;
we call this the use of ``batch-negatives''.
The same binary focal loss is used for training.
As will be shown in Section~\ref{sec:selft:res} and is commonly observed
in the self-supervised literature~\cite{oord2018representation,chen2020simple,arandjelovic2017look,he2020momentum},
batch-negatives are crucial to obtain good performance.

\paragraph{Relation to other methods.}
While multiple works have used self-training with pseudo-labelling to boost
the detector performance, none
follow the above approach.
Pseudo-labelling is popular in the weakly-supervised
(classes in the image are specified but not their location)
or semi-supervised
low-shot works~\cite{redmon2017yolo9000,sohn2020simple,zoph2020rethinking,ramanathan2020dlwl}
with closed vocabulary detectors.
This means that pseudo-labelling is easier as all classes are seen during training,
but also that the self-training follows exactly the same setup as the initial training,
where the negatives are other classes and there is no need or way to use
batch-negatives.

Our \emph{\threerec} uses the pseudo-detections from the \emph{\tworec} detector
conditioned on the image caption, while
Detic~\cite{zhou2022detecting} computes the pseudo-detection independently
of the caption by taking the largest bounding box proposal.
Detic adopts batch-negatives but does so on the image-level rather than the bounding box-level;
a more detailed discussion is available in \isArXiv{Appendix~\ref{sec:app:selft}.}{the supplementary material.}
GLIPv2~\cite{zhang2022glipv2} also uses batch-negatives,
but does not consider the zero-shot scenario.
While we use the entire caption at once to produce pseudo-detections,
GLIPv2 extracts noun phrases and pseudo-labels them individually.
This could provide better quality pseudo-labels, but comes with its downsides as well,
in that it depends on the quality of the text parser and requires additional bookkeeping
and special handling of repeated noun phrases in the batch.

\vspace*{-0.15cm}%
\subsubsection{Results}
\label{sec:selft:res}
\vspace{-0.2cm}

\tabSelfTrain
\tabSota

\paragraph{Implementation details.}
We start from the strong \emph{\tworec} detector from Section~\ref{sec:init}
and verify that longer training on \LVISminusrare does not improve
results further.
Conceptual Captions 12M~\cite{changpinyo2021conceptual} (CC12M), an image-caption dataset gathered
automatically from the internet containing 12M images,
is used for all self-training experiments.
The self-training starts from the \emph{\tworec} detector checkpoint
and continues training, where each training step simultaneously optimizes
the losses on a batch of \LVISminusrare images with true ground-truth
and a batch of pseudo-labelled images from CC12M.
In line with~\cite{zhou2022detecting}, we find we can reduce the resolution
of the CC12M images (for \LVISminusrare we use $800 \times 1024$,
while for CC12M $400 \times 512$ is sufficient)
thus fitting a larger number of images in the batch and allowing for
more batch-negatives.

\paragraph{Performance.}
Table~\ref{tab:selftrain} shows the self-training results --
it is clear that our self-training, \emph{\threerec}, significantly improves
both metrics and on both backbones, providing an especially large boost
for the unseen classes.

\paragraph{Comparison to Detic~\cite{zhou2022detecting}.}
We do not simply copy the numbers from~\cite{zhou2022detecting}
as this wouldn't be a fair comparison -- we use different visual backbones,
detector type, the self-training dataset, \etc.
Furthermore, Detic~\cite{zhou2022detecting} focuses on the zero-shot and not
open-vocabulary aspect (\eg the full approach specifically searches for
the LVIS-rare classes in the captions and uses this as the pseudo-label).
Therefore, we reimplement the open-vocabulary version of Detic$^\dagger$,
using our \emph{\tworec} detector (see \isArXiv{Appendix~\ref{sec:app:selft}}{the supplementary material for details}).
Detic$^\dagger$ performs well, giving improvements over \emph{\tworec}.
However, our self-training approach also significantly
beats Detic$^\dagger$.
We also compare to another approach proposed by~\cite{zhou2022detecting}
where the pseudo-detection is simply taken to be the image bounding box.
In fact, \emph{Image bbox} beats Detic$^\dagger$ slightly,
but \emph{\threerec} is still superior.

\paragraph{Importance of batch-negatives.}
As an ablation, we also train versions of
the \emph{Image bbox} and our \emph{\threerec} approaches where batch-negatives
are not used. The results show a clear large benefit of using
batch-negatives -- without them there is barely any gain from self-training
as the task is too easy.

\section{Results and discussion}
\label{sec:res}

Comparison on an equal footing with the state-of-the-art is hard
because most works use different visual backbones,
pretraining, detector architecture, training procedure,
augmentations, \etc.
Sections~\ref{sec:aug:res},~\ref{sec:init:res} and~\ref{sec:selft:res}
demonstrate the performance of each of our methods individually
through fair comparisons where all these aspects are identical,
while here we resort to the standard practice of reporting system-level
performance (Table~\ref{tab:sota}).
We list best performing methods that are truly zero-shot and open vocabulary,
and are trained following the
\LVISminusrare benchmark rules (\ie the only detection annotations used are
the LVIS training set with the rare classes removed).
For example, this criterion disqualifies
PromptDet~\cite{feng2022promptdet} and
the best performing versions of Detic~\cite{zhou2022detecting}
(they actively use the list of LVIS classes to pseudo-label additional images,
\ie not open vocabulary),
FIBER~\cite{dou2022coarsetofine} and some OWL~\cite{minderer2022simple} experiments
(train on many more detection annotations),
GLIP~\cite{li2022glip} and GLIPv2~\cite{zhang2022glipv2}
(rare classes are not removed during training so not zero-shot, and
more training data is used),
\etc.

Our final open vocabulary detector, \emph{\threerec}, achieves the highest
\APall and competitive \APrare.
On \APall it sets the state-of-the-art by a large margin
-- the largest network (NFNet-F6 with 440M parameters)
achieves 44.6\% (and 43.5\% without self-training) while the best second
is at 35.3\% (OWL~\cite{minderer2022simple}'s VIT-H/14 with 630M parameters),
making for an impressive improvement of 9.3\% points
(8.2\% without self-training).
Even our smaller network (NFNet-F0 with 71M parameters)
with self-training beats the previously best reported performance.

On the unseen classes, \APrare, we compare favourably to the latest
approaches. Only the concurrent \isArXiv{}{unpublished} F-VLM~\cite{kuo2023fvlm} method
performs better,
achieving 32.8\% for the largest R50x64 model,
while our equally large NFNet-F6 is a close second at 30.1\%.
It should be noted that, when trained on full LVIS, it has been observed
that \APrare inherently has high variance as these are the long tail categories,
and an absolute difference of 1\% might not be
significant~\cite{lvisbestpractices};
for example, our best performing run achieves 31.7\%.
The shared third place with a significantly lower \APrare value of 25.6\%
is achieved by OWL with VIT-L/14 (310M parameters)
and our much smaller NFNet-F0 model (71M parameters).

The good performance is partially due to the use of the strong vision backbone.
However, simply using it out of the box (\emph{\zerorec}) fails
(Table~\ref{tab:sota}, \cf Section~\ref{sec:aug:res})
and our methods are required to unlock its power.

\tabTransfer

\paragraph{Transfer}
capabilities are tested by evaluating the
\LVISminusrare trained networks on COCO~\cite{lin2014coco} and Objects365-v1~\cite{shao2019objects365}.
Table~\ref{tab:transfer}
shows the networks achieve impressive performance:
on COCO
even our smallest model without self-training beats all previous approaches,
while on Objects365 (estimated by~\cite{kuo2023fvlm} to have only 63\% overlap  %
with \LVISminusrare training classes)
\emph{\threerec} improves upon the previous best mAP by 5.1\% points.

Qualitative results are provided in \isArXiv{Appendix~\ref{sec:app:qual}.}{the supplementary.}

\section{Conclusions}

We introduced three methods for improving alignment between
visual and text features, which in turn boosts zero-shot
detection performance.
They reduce overfitting and forgetting of concepts learnt during
pretraining, improve training speed while decreasing accelerator memory
requirements, and make use of large image-text datasets without
costly detection annotations.
We achieve superior \APall on the challenging \LVISminusrare benchmark,
and transfer to COCO and Objects365, while obtaining \APrare competitive
with concurrent \isArXiv{}{unpublished} work.
Further research directions include investigating how to even more
efficiently make use of plentiful image-text data with
improved pseudo-labelling, losses, or combinations with
self-supervised learning.

\isArXiv{
\paragraph{Acknowledgments.}
We thank Evan Shelhamer for fruitful discussions
and Iain Barr for help with the codebase.
}{}

{\small
\bibliographystyle{ieee_fullname}
\bibliography{bib/shortstrings,bib/more,bib/vgg_other,bib/vgg_local}
}

\isArXiv{
\appendix
\ifdefined\isArXiv
\else
\documentclass[10pt,twocolumn,letterpaper]{article}

\usepackage{iccv}
\usepackage{times}
\usepackage{epsfig}
\usepackage{graphicx}
\usepackage{amsmath}
\usepackage{amssymb}

\usepackage{subcaption}
\usepackage{tabularx}
\usepackage[usenames,dvipsnames]{xcolor}
\usepackage{booktabs}  %
\usepackage{fancyvrb}
\usepackage{fvextra}
\makeatletter
\@namedef{ver@everyshi.sty}{}
\makeatother
\usepackage{tikz}
\usetikzlibrary{shapes}

\usepackage[numbers,sort]{natbib}

\usepackage{xspace}
\def\eg{\emph{e.g.}\@\xspace}
\def\ie{\emph{i.e.}\@\xspace}
\def\etal{\emph{et al.}\@\xspace}
\def\cf{\emph{c.f.}\@\xspace}
\def\etc{\emph{etc}}

\renewcommand{\paragraph}[1]{\smallskip\noindent{\bf{#1}}}

\def\APall{mAP\textsubscript{all}\@\xspace}
\def\APrare{mAP\textsubscript{rare}\@\xspace}
\def\APcomm{mAP\textsubscript{comm}\@\xspace}
\def\APfreq{mAP\textsubscript{freq}\@\xspace}
\def\LVISminusrare{LVIS\textsubscript{-R}\@\xspace}

\def\zerorec{0Ways\@\xspace}
\def\onerec{1Ways\@\xspace}
\def\tworec{2Ways\@\xspace}
\def\threerec{3Ways\@\xspace}

\newcommand{\isArXiv}[2]{#1}
\usepackage[pagebackref=true,breaklinks=true,letterpaper=true,colorlinks,bookmarks=false]{hyperref}

\def\iccvPaperID{1802}
\def\httilde{\mbox{\tt\raisebox{-.5ex}{\symbol{126}}}}

\ificcvfinal\pagestyle{empty}\fi

\begin{document}

\def\fulltitle{Three ways to improve feature alignment for open vocabulary detection}
\title{\fulltitle \\ -- Supplementary material --}
\hypersetup{pdfauthor={\iccvPaperID},pdftitle={\iccvPaperID: \fulltitle}}

\maketitle
\ificcvfinal\thispagestyle{empty}\fi

\tableofcontents
\fi

\newcommand{\tabArhParams}{
\def\y{\checkmark}
\def\n{~}
\begin{table*}[t]
\centering
\begin{tabular}{ccccrr} \toprule
&&&& \multicolumn{2}{c}{Batch size} \\
\cmidrule{5-6}
Backbone & Self-training & Init lr & Device & \LVISminusrare & CC12M \\
\midrule
NFNet-F0 & \n & $3 \times 10^{-4}$ & 16 TPUv3 chips & 128 & / \\
NFNet-F0 & \y & $3 \times 10^{-4}$ & 16 TPUv3 chips & 128 & 1024 \\
NFNet-F6 & \n & $1 \times 10^{-4}$ & 16 TPUv3 chips & 64 & / \\
NFNet-F6 & \y & $1 \times 10^{-4}$ & 32 TPUv4 chips & 32 & 256 \\
\bottomrule
\end{tabular}
\caption{{\bf Architecture-specific optimization parameters.}
}
\label{tab:archparams}
\end{table*}
}

\newcommand{\tabLvisAll}{
\begin{table*}[t]
\definecolor{cheat}{HTML}{999999}
\centering
\begin{tabular}{ccccccc} \toprule
& & & \multicolumn{2}{c}{LVIS} & \multicolumn{2}{c}{Transfer} \\
\cmidrule(l){4-5} \cmidrule(l){6-7}
Method & Train dataset & Backbone & \APall & \APrare & COCO & O365 \\
\midrule
\tworec & LVIS & NFNet-F0 & \textcolor{cheat}{\meanstd{34.4}{0.11}} & \textcolor{cheat}{\meanstd{23.7}{1.41}} & \meanstd{40.8}{0.11} & \meanstd{14.7}{0.07} \\
\tworec & \LVISminusrare & NFNet-F0 & \meanstd{33.8}{0.15} & \meanstd{20.9}{0.34} & \meanstd{40.6}{0.14} & \meanstd{14.6}{0.04} \\
\threerec & \LVISminusrare & NFNet-F0 & \meanstd{35.7}{0.20} & \meanstd{25.6}{1.12} & \meanstd{41.5}{0.16} & \meanstd{16.4}{0.15} \\
\tworec & LVIS & NFNet-F6 & \textcolor{cheat}{\meanstd{45.1}{0.18}} & \textcolor{cheat}{\meanstd{37.0}{0.97}} & \meanstd{46.5}{0.04} & \meanstd{20.8}{0.05} \\
\tworec & \LVISminusrare & NNetF-F6 & \meanstd{43.5}{0.12} & \meanstd{27.6}{0.80} & \meanstd{46.5}{0.26} & \meanstd{20.3}{0.05} \\
\threerec & \LVISminusrare & NFNet-F6 & \meanstd{44.6}{0.31} & \meanstd{30.1}{1.83} & \best{\meanstd{46.9}{0.05}} & \best{\meanstd{22.8}{0.14}} \\
\bottomrule
\end{tabular}
\caption{{\bf Training on all of LVIS vs self-training.}
Training \emph{\tworec} on full LVIS predictably performs better on LVIS than when
training on \LVISminusrare (\ie LVIS without annotations for rare classes).
However, it loses to the self-trained detector (\emph{\threerec}) when transferring to COCO and Objects365.
}
\label{tab:lvisall}
\end{table*}
}

\newcommand{\figOrig}{
\begin{figure*}[t]
\definecolor{arrow}{HTML}{6aa84f}
\definecolor{gate}{HTML}{ff9900}
\definecolor{back}{HTML}{cfe2f3}
\includegraphics[width=\linewidth]{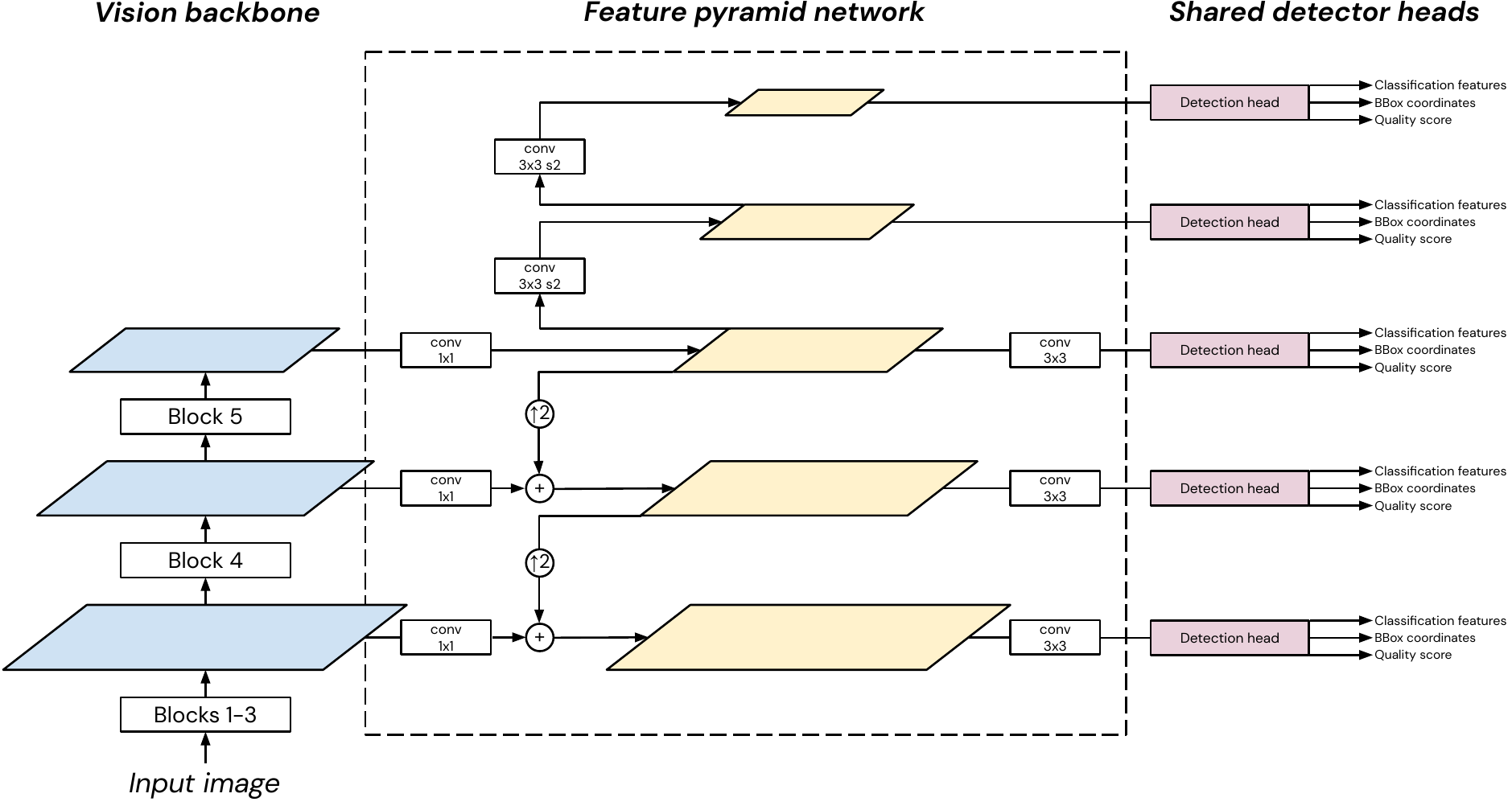}
\caption{{\bf Standard single-stage detector architecture.}
\textcolor{CornflowerBlue}{Light blue} and \textcolor{Apricot}{light yellow} parallelograms
represent the backbone and FPN feature maps, respectively;
circles with ↑2 are non-trainable up--sampling,
squares are trainable modules (\eg convolutions),
convolution blocks show the kernel size
and potential striding (`s2': stride 2).
The alignment preserving architecture is shown in \isArXiv{Figure~\ref{fig:gating}.}{Figure 3 of the main paper.}
}
\label{fig:orig}
\end{figure*}
}

\newcommand{\zs}[1]{\textcolor{RedOrange}{#1}}
\newcommand{\ms}[1]{\textcolor{Gray}{#1}}

\newcommand{\figQualThreeVsTwo}{
\begin{figure*}[t]
\def\imW{0.495\linewidth}
\centering%
\begin{tabular}{c@{~~}c}
\includegraphics[width=\imW]{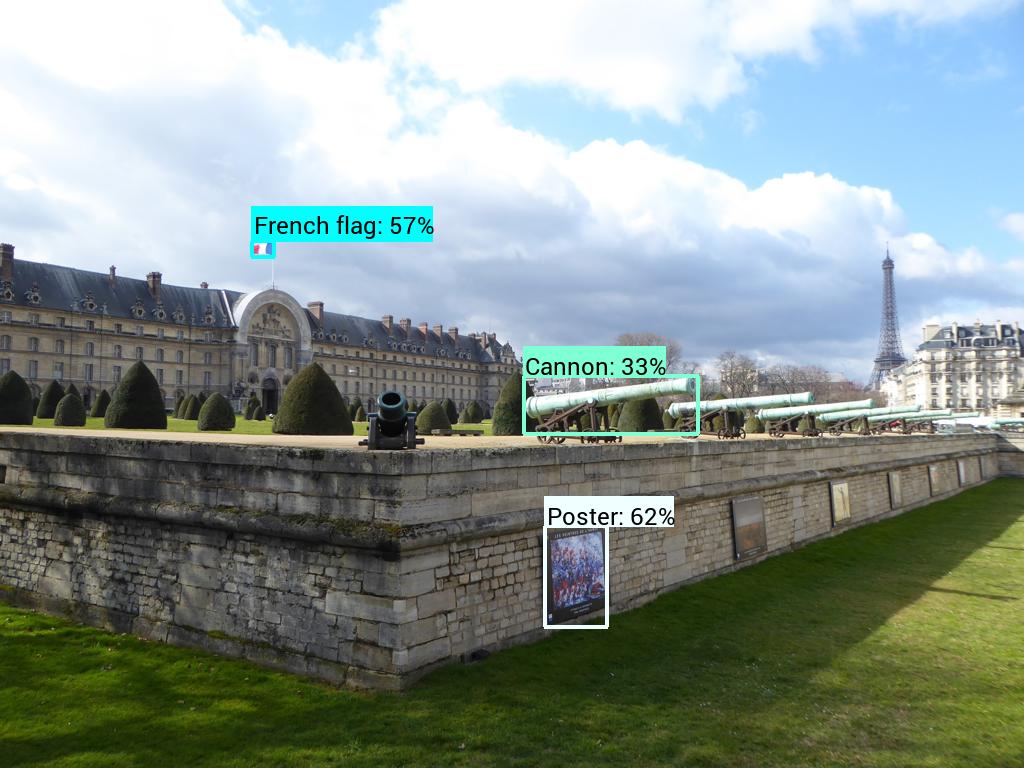}
&
\includegraphics[width=\imW]{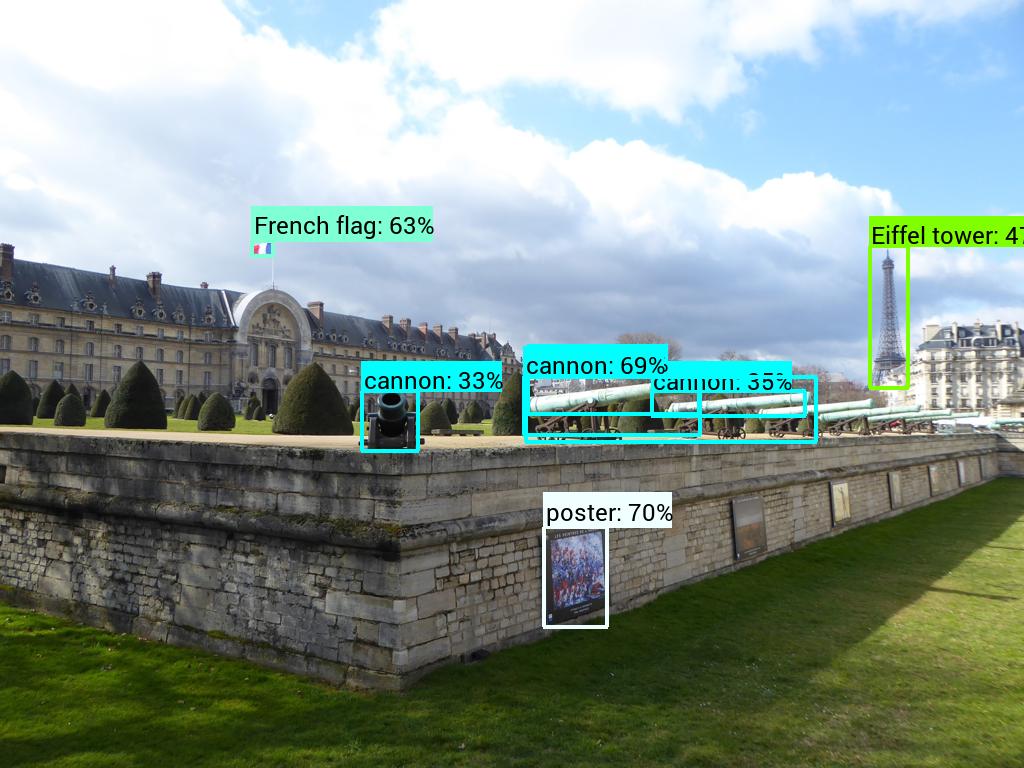}
\\
\includegraphics[width=\imW]{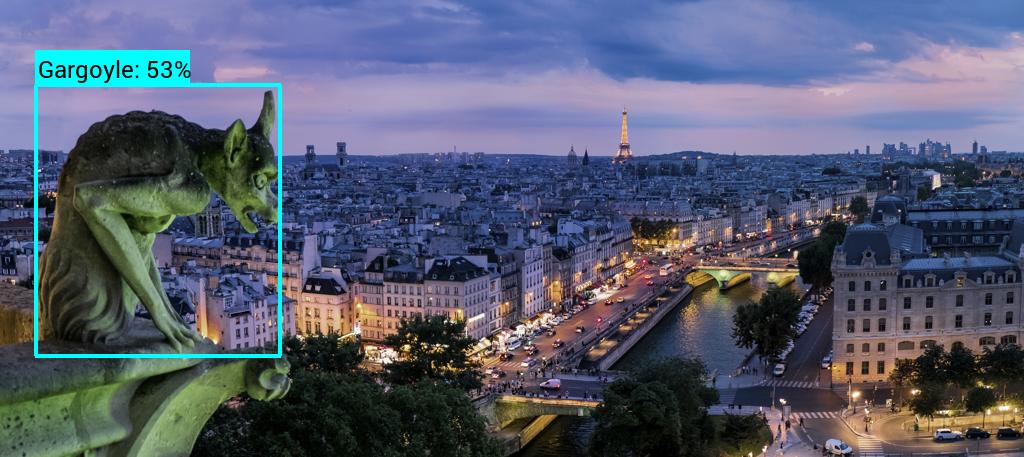}
&
\includegraphics[width=\imW]{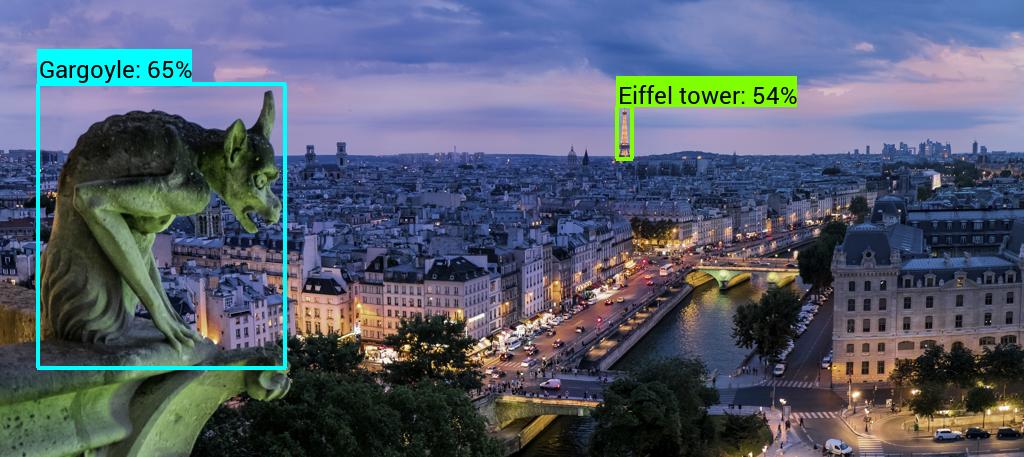}
\\
\includegraphics[width=\imW]{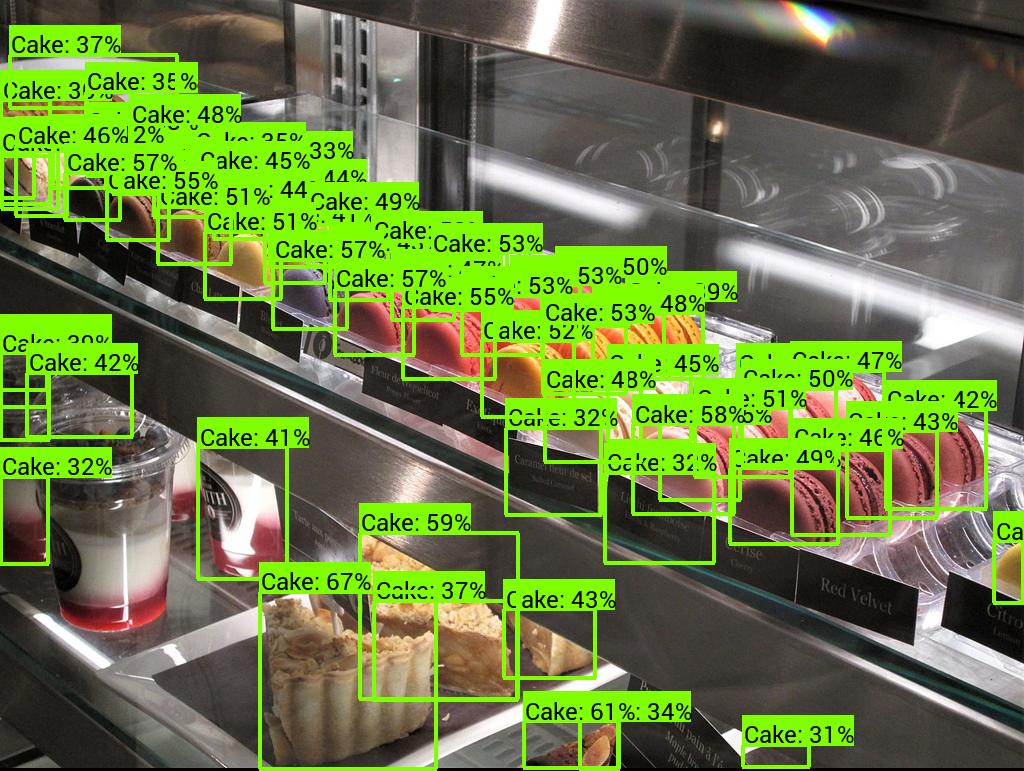}
&
\includegraphics[width=\imW]{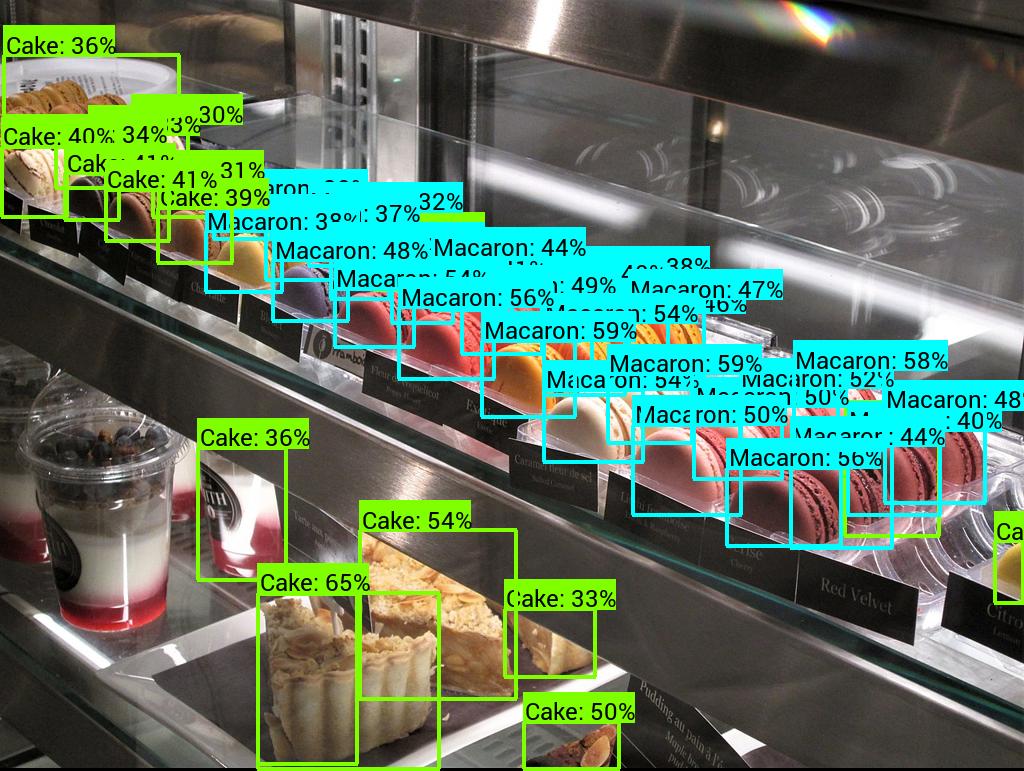}
\\
(a) \tworec & (b) \threerec
\end{tabular}
\caption{{\bf \emph{\tworec} vs \emph{\threerec} (NFNet-F6).}
Detections with a score larger than 0.3 are shown.
The queries for the three examples are:
(top) \zs{Eiffel tower}, \zs{French flag}, \zs{cannon}, \ms{poster};
(middle) \zs{Eiffel tower}, \zs{gargoyle};
(bottom) \ms{cake}, \zs{macaron}.
No human-annotated boxes were seen during training for
\zs{all queries} apart from \ms{poster} and \ms{cake}.
\emph{\tworec} has zero-shot detection capability,
but \emph{\threerec} generally outperforms it.
}
\label{fig:qual:threevstwo}
\end{figure*}
}

\newcommand{\figQualThreeParis}{
\begin{figure*}[t]
\def\imHu{4.7cm}
\def\imHd{4.79cm}
\includegraphics[height=\imHu]{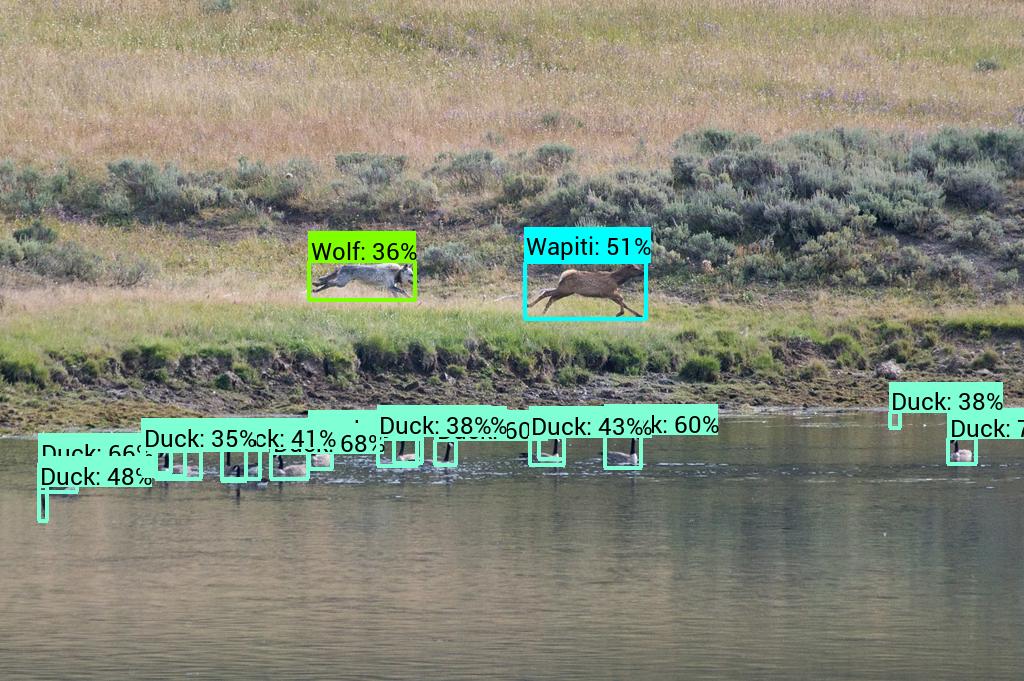}
\includegraphics[height=\imHu]{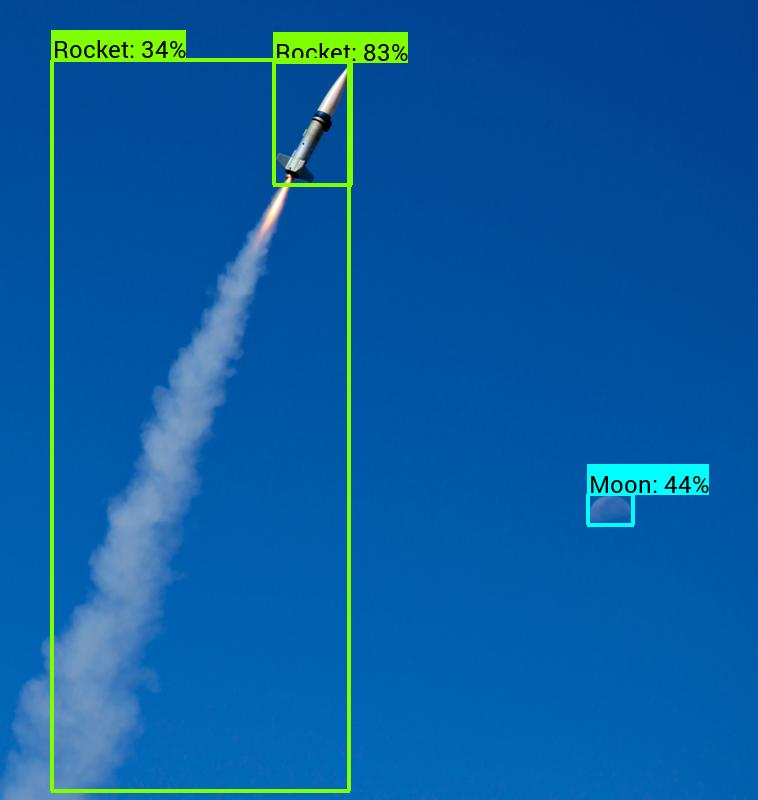}
\includegraphics[height=\imHu]{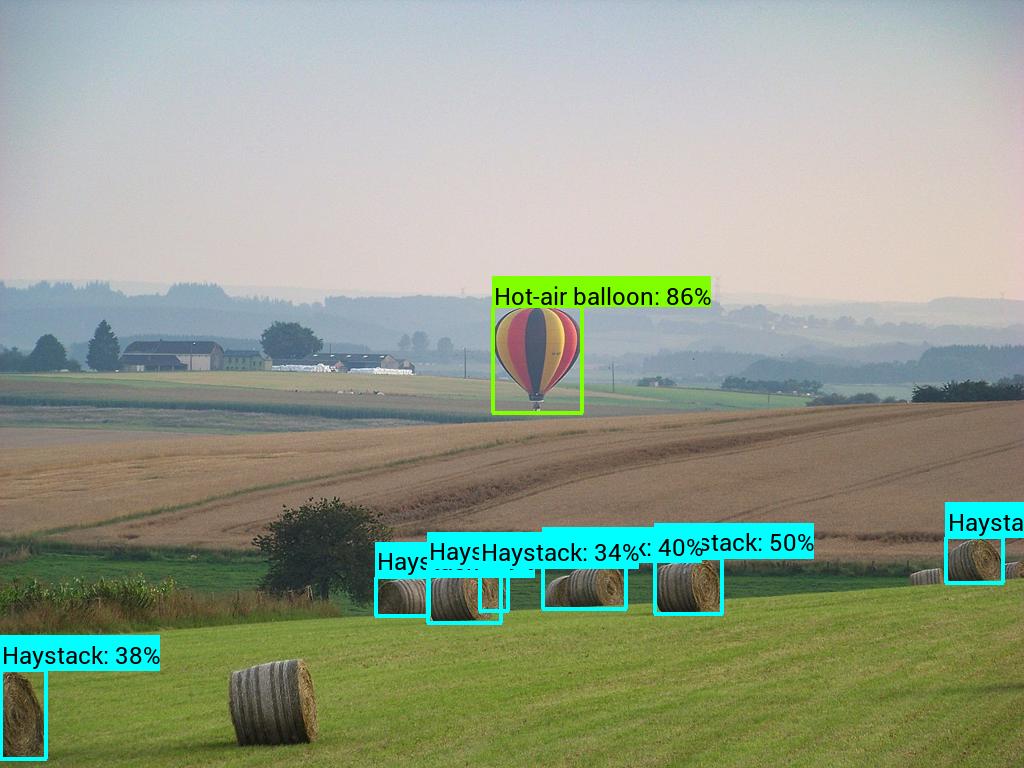} \\
\includegraphics[height=\imHd]{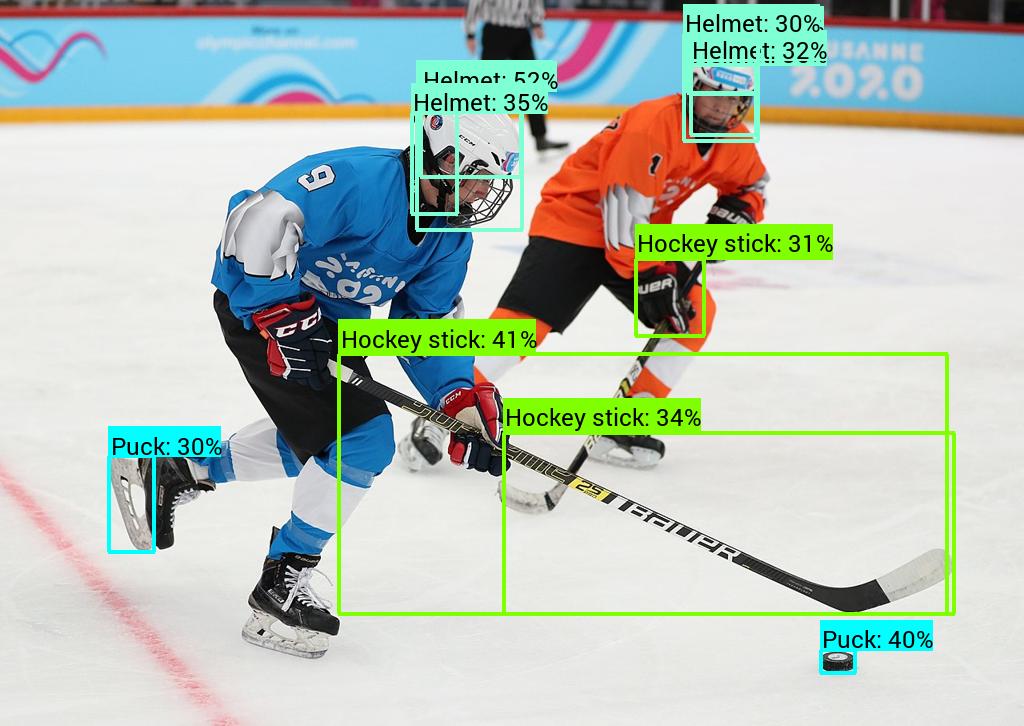}
\includegraphics[height=\imHd]{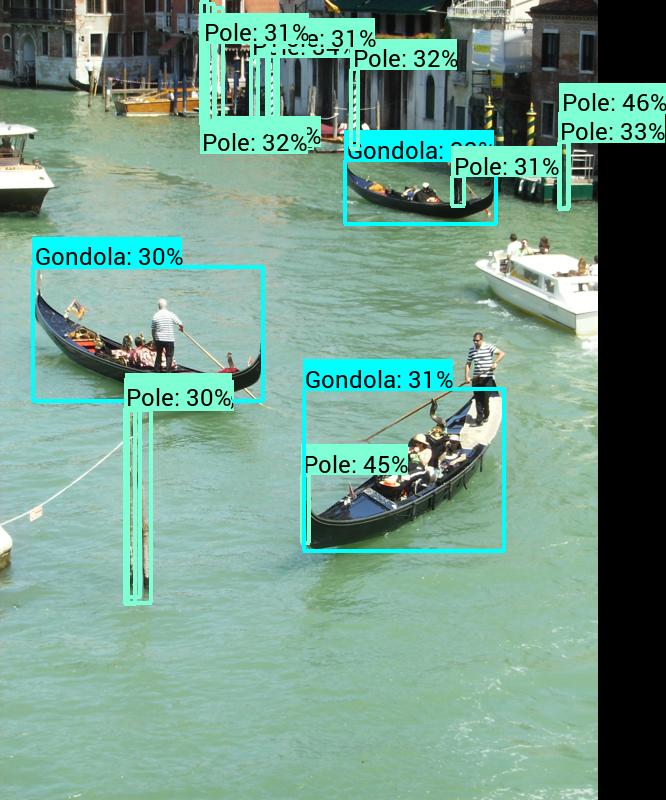}
\includegraphics[height=\imHd]{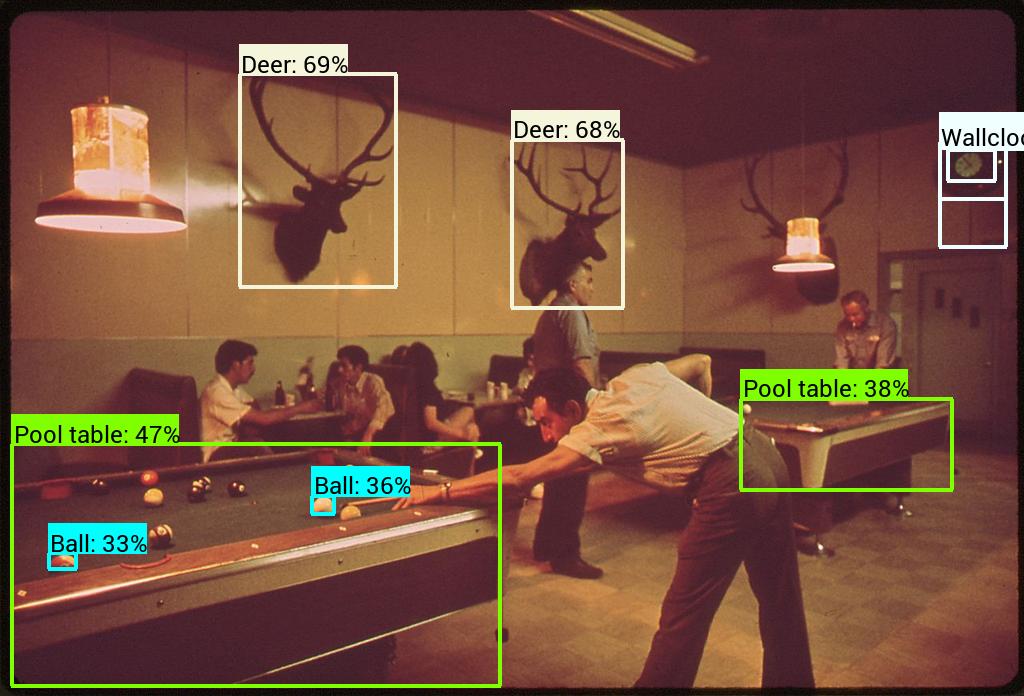}
\caption{{\bf \emph{\threerec} (NFNet-F6) zero-shot detection examples.}
Detections with a score larger than 0.3 are shown.
The queries for the six examples are:
(top-left) \zs{wolf}, \zs{wapiti}, \ms{duck};
(top-middle) \zs{rocket}, \zs{moon};
(top-right) \zs{hot-air balloon}, \zs{haystack};
(bottom-left) \ms{helmet}, \zs{puck}, \zs{hockey stick};
(bottom-middle) \zs{gondola}, \ms{pole};
(bottom-right) \zs{pool table}, \ms{ball}, \ms{deer}, \ms{wallclock}.
Networks were trained with human-annotated boxes for
\ms{duck}, \ms{helmet}, \ms{pole}, \ms{ball}, \ms{deer} and \ms{wallclock},
while no human-annotated boxes were seen during training for
\zs{wolf}, \zs{wapiti}, \zs{rocket}, \zs{moon}, \zs{hot-air balloon}, \zs{haystack}, \zs{puck}, \zs{hockey stick},
\zs{gondola}, and \zs{pool table}.
Note that \ms{dog} (a frequent training category) rightfully has a
lower confidence than \zs{wolf} for the top-left image.
}
\label{fig:qual:threeparis}
\end{figure*}
}

\newcommand{\figQualLvis}{
\begin{figure*}[t]
\vspace{-0.5cm}
\centering
\includegraphics[width=0.95\linewidth]{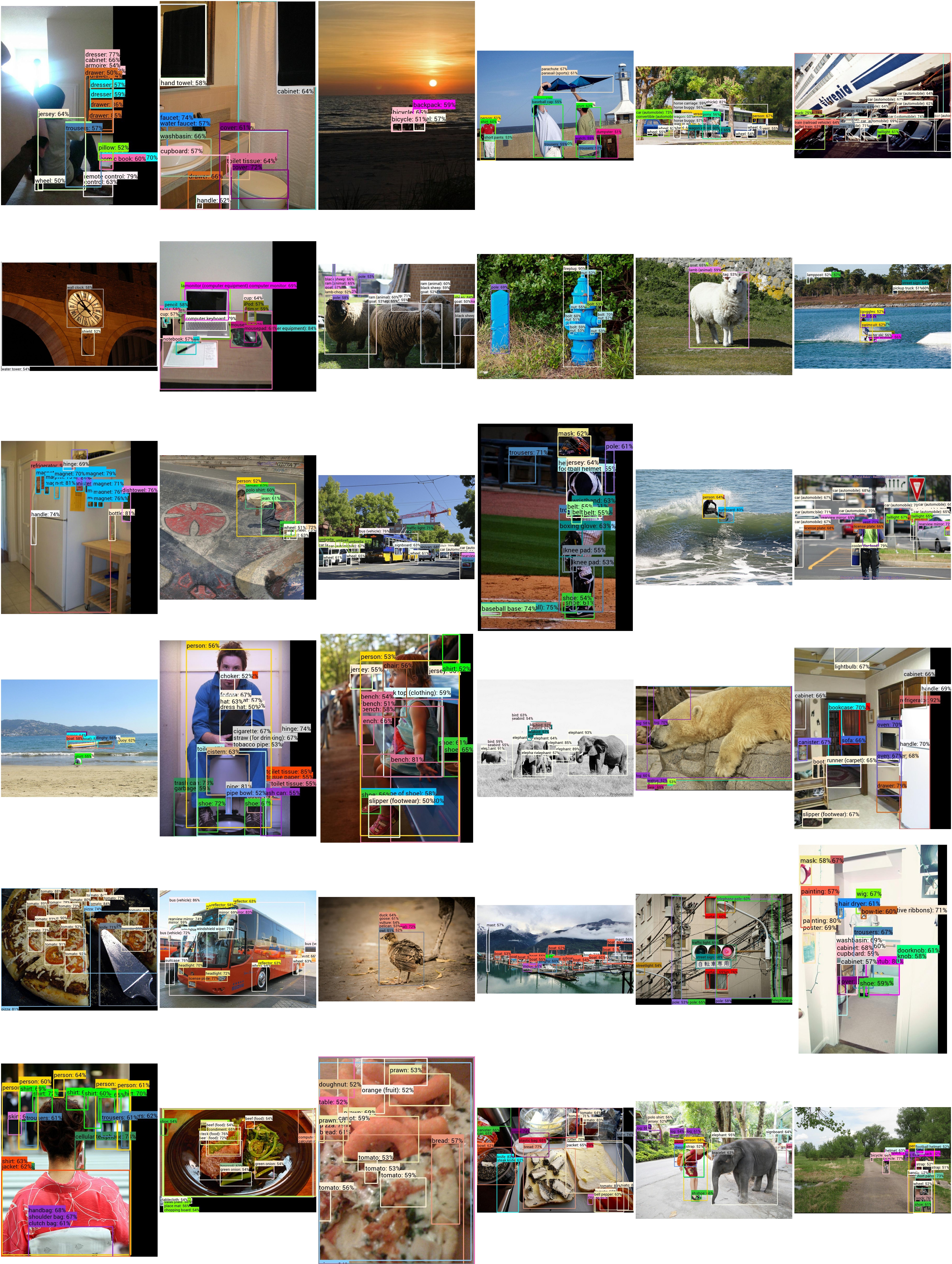}
\caption{{\bf Random sample: LVIS validation set (\emph{\threerec}, NFNet-F6).}
Up to 20 detections with a score larger than 0.5 are shown.
Best viewed digitally.
}
\label{fig:qual:lvis}
\end{figure*}
}

\newcommand{\figQualCoco}{
\begin{figure*}[t]
\vspace{-0.5cm}
\centering
\includegraphics[width=0.95\linewidth]{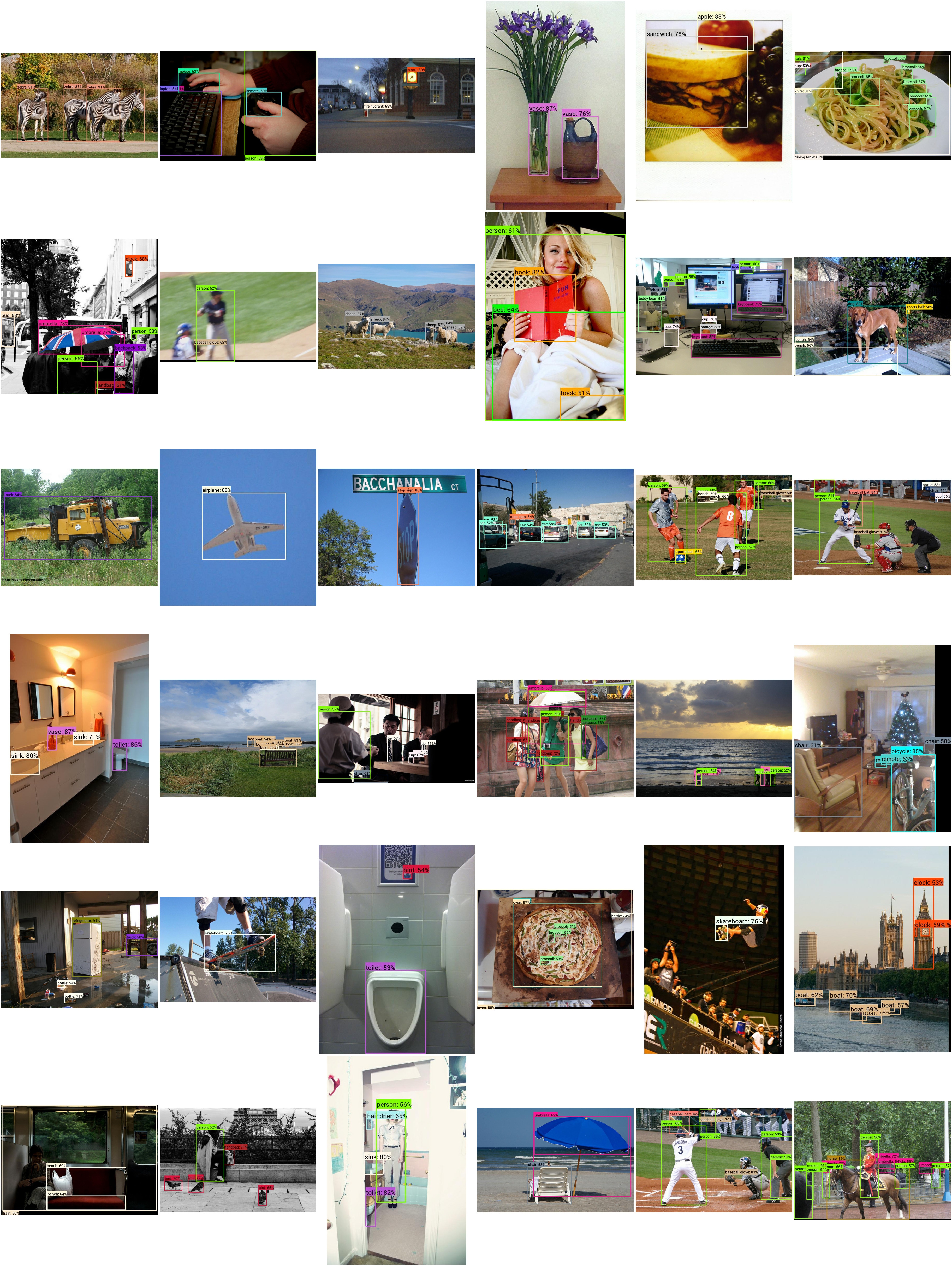}
\caption{{\bf Random sample: COCO validation set (\emph{\threerec}, NFNet-F6).}
The detector is trained on \LVISminusrare and transferred
to COCO without any additional training.
Up to 20 detections with a score larger than 0.5 are shown.
Best viewed digitally.
}
\label{fig:qual:coco}
\end{figure*}
}

\newcommand{\figQualObj}{
\begin{figure*}[t]
\vspace{-0.5cm}
\centering
\includegraphics[width=0.95\linewidth]{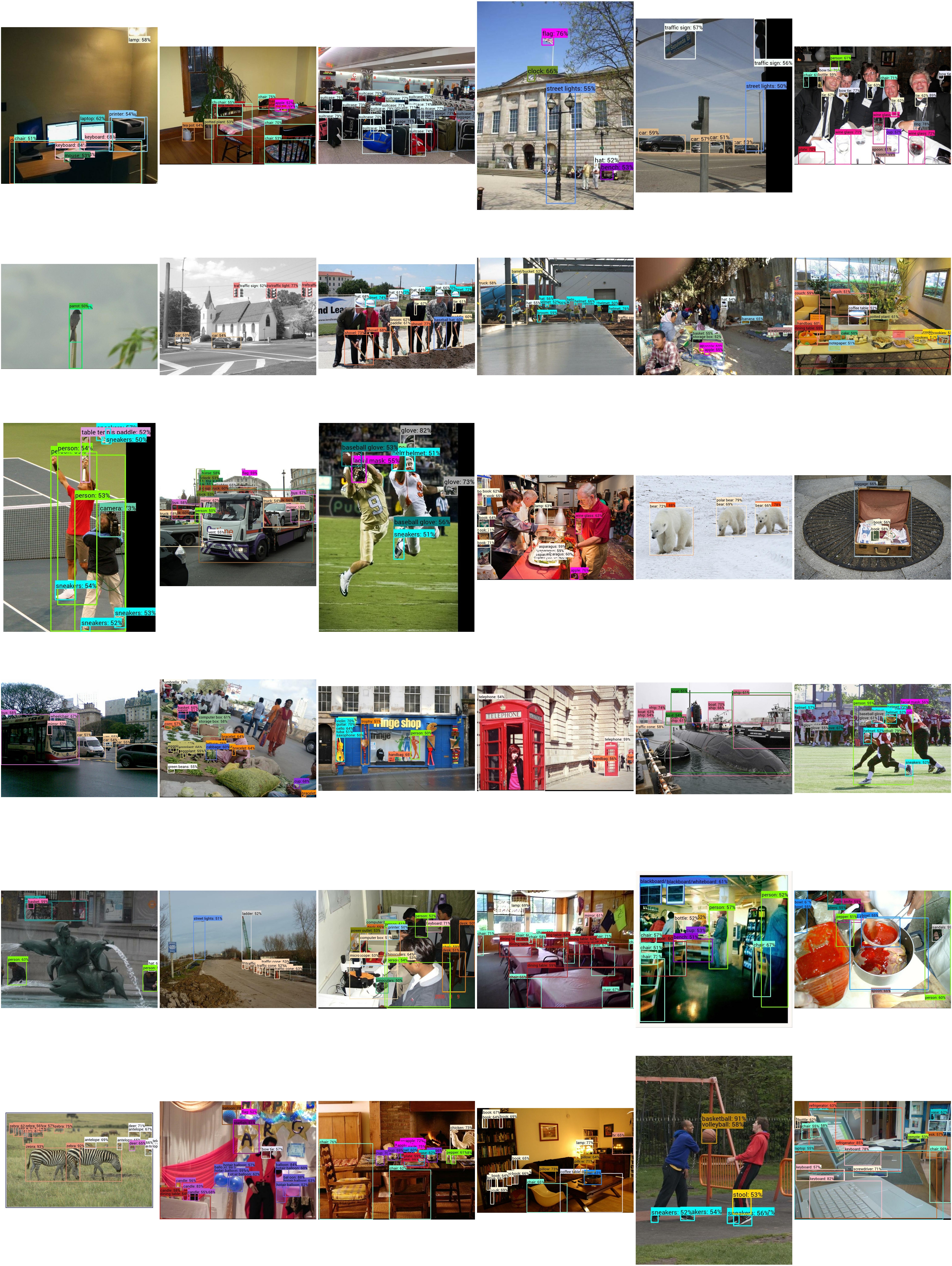}
\caption{{\bf Random sample: Objects365-v1 validation set (\emph{\threerec}, NFNet-F6).}
The detector is trained on \LVISminusrare and transferred
to Objects365 without any additional training.
Up to 20 detections with a score larger than 0.5 are shown.
Best viewed digitally.
}
\label{fig:qual:o365}
\end{figure*}
}

\section*{Appendix overview}

Appendix~\ref{sec:app:imp} contains full details of the detector architecture
and training procedure,
while
Appendix~\ref{sec:app:init} provides more information on the
alignment preserving architecture.
In Appendix~\ref{sec:app:selft} we give further details on the
implementation of self-training, reimplementation of Detic, and
relation to other methods.
Finally, qualitative detection examples are shown in Appendix~\ref{sec:app:qual}.

\section{Implementation details}
\label{sec:app:imp}

\paragraph{Architecture.}
For closed vocabulary detectors, the final fully connected layer
also trains a per-class bias, and initializing it according to the focal loss
scheme~\cite{lin2017focal} stabilizes the beginning of training.
For the open vocabulary case it is not possible to have this per-class bias term,
so we use a learnt shared bias, \ie a single scalar added to the scalar product
between the classification features and the text embeddings,
initialized in the same way, \ie
$\text{bias}_\text{init} = -\log\left(\left(1 - p\right) / p\right)$, where $p=0.01$.
In initial experiments we observed this indeed aided stability.

The original T-Head detector head architecture~\cite{feng2021tood} uses
the ``task-aligned predictor'' (TAP) only on the regression and classification branches
and not on the quality branch. For consistency we simply use TAP
on all three branches.

\paragraph{Input and annotation processing.}
Training and evaluation is done with $800 \times 1024$ images for all datasets
(\LVISminusrare, COCO, Objects365) apart from the pseudo-labelled images
from CC12M where we found that $400 \times 512$ is sufficient.
Training uses large scale jittering~\cite{ghiasi2021simple}
for \LVISminusrare and small scale jittering for CC12M.
Following~\cite{zhou2021probabilistic,minderer2022simple},
50 pseudo-negatives are used for each training sample.

\paragraph{Total loss} is the sum of three equally weighted standard losses:
binary focal loss~\cite{lin2017focal} with default FCOS parameters ($\gamma=2$, $\alpha=0.25$),
gIoU loss\cite{rezatofighi2019generalized},
and the IoU prediction loss~\cite{wu2020iou}.

\tabArhParams

\paragraph{Optimization} is done with AdamW~\cite{loshchilov2019decoupled} and $10^{-4}$ weight decay,
and,
as is standard for NFNets,
adaptive gradient clipping~\cite{brock2021highperformance}
with the clipping threshold of 0.04 is used.
Depending on the backbone architecture and whether self-training is performed,
we adjust the training hardware, batch size and initial (post-warmup) learning rate
-- Table~\ref{tab:archparams} shows the parameters for all configurations.
All networks are trained for 32 epochs
and the learning rate schedule consists of linear warm-up for 0.25 epochs,
followed by 10 fold reductions at 67\% and 89\% of the training.
The same schedule is used for self-training.

\isArXiv{The implementation uses JAX~\cite{google2018jax} and the DeepMind JAX ecosystem~\cite{deepmind2020jax}.}{}

\section{Alignment preserving architecture}
\label{sec:app:init}

\figOrig

\isArXiv{Figure~\ref{fig:gating}}{Figure~3 of the main paper} shows our alignment preserving architecture
with shortcuts and trainable gates.
For reference, Figure~\ref{fig:orig} shows the classic
design~\cite{lin2017feature,tian2019fcos} of a single-stage detector.

\paragraph{Trainable gating.}
Recall that the gates perform the following operation:
$o = x (1-\tan\alpha) + y \tan{\alpha}$,
where $x$ and $y$ are the inputs and $\alpha$ is a learnt parameter initialized to $0$.
We briefly experimented with the Flamingo~\cite{alayrac2022flamingo} formulation:
$o = x + y \tan{\alpha}$, and found it performs similarly.

\paragraph{Matching dimensionalities.}
For clarity of presentation,
we glanced over one detail that we elaborate on here --
gates and shortcuts assume the dimensionalities of all input and output
features are the same,
while the default design of the feature pyramid network (FPN)
and detection heads includes changes in the channel dimension.
We do not attempt to fix this by changing the default settings for the FPN and detector head
dimensionalities and making all channels the same size,
as that would increase the parameter count, be less memory and computation efficient,
and be harder to compare fairly with existing approaches.

Specifically, the issue is the initial $1 \times 1$ convolution in the FPN
applied on the final backbone features, as this reduces channels to 256
and thus loses information.
All following processing maintains 256-D features until the very end
where there is an increase back to the original dimensionality
to produce classification features compatible with text embeddings
(786-D and 1376-D for NFNet-F0 and NFNet-F6, respectively).
We address this pragmatically -- the dimensionality reduction convolution is initialized
such that distances and scalar products are preserved as much as possible,
though use of an orthogonal projection.
Similarly, the dimensionality increase at the end of the network
is initialized to the transpose of this projection.
We find that the details are not crucial -- in fact the axis aligned
orthogonal matrix (\ie a rectangular ``cropped'' identity matrix)
is used for simplicity, and even without this initialization
the training still works well as it only needs to learn one
projection to bring features back into alignment.

\section{Self-training}
\label{sec:app:selft}

\paragraph{Implementation details.}
Self-training simultaneously processes a batch from \LVISminusrare
with human-annotations and a batch from CC12M with pseudo-annotations.
All the three losses (classification, bounding box regression, quality prediction)
are used for the pseudo-labelled data as well as this performs slightly better
than only using the classification loss.
For Detic$^\dagger$ and \emph{Image bbox} baselines we found that,
consistent with the paper~\cite{zhou2022detecting},
using only the classification loss performs best, and report only these results.
The losses for the two human- and pseudo-labelled data are simply summed up,
different relative weighting was not found to be beneficial for any of
the methods or baselines.

\paragraph{Relation to other methods.}
We elaborate on the comparison to other methods from \isArXiv{Section~\ref{sec:selft}}{Section~3.3 of the main paper}.
Recall that~\cite{redmon2017yolo9000,sohn2020simple,zoph2020rethinking,ramanathan2020dlwl}
are closed vocabulary detectors and and there is no need or way to use
batch-negatives.
GLIPv2~\cite{zhang2022glipv2} does not consider the zero-shot scenario
and uses noun phrases instead of entire captions like we do.
Furthermore, their deep fusion architecture restricts the use
of batch-negatives to the pre-fusion features making its effect
indirect, while our loss operates on the very final detector head
classification features which are directly used for detecting the object.

In contrast to our approach and GLIPv2,
Detic~\cite{zhou2022detecting} does not make use of the image caption
to produce the pseudo-detection, but instead
picks the largest bounding box proposal instead. The method does use
batch-negatives in an additional loss,
although this is done on the global image-level rather
than the bounding box-level.

Finally, the Detic paper~\cite{zhou2022detecting} also implements
baselines where the closed vocabulary pseudo-labelling methods mentioned above~\cite{redmon2017yolo9000,sohn2020simple,zoph2020rethinking,ramanathan2020dlwl}
are adapted to the open vocabulary setting.
For these baselines, the paper uses the image caption to produce pseudo-detections,
but does not use batch-negatives.
We showed in \isArXiv{Section~\ref{sec:selft:res}}{Section~3.3.1 of the main paper} that batch-negatives are crucial for
good performance.

\paragraph{Comparisons with Detic~\cite{zhou2022detecting}.}
Detic~\cite{zhou2022detecting} focuses on the zero-shot detection task
but the main results forgo the open vocabulary claim as
the knowledge of the names of the \LVISminusrare test classes
is explicitly used during training,
\eg the captions are filtered for LVIS class names and this noisy label is
assigned to the image.
We focus on zero-shot open vocabulary detection and do not use test class names
during training, so we compare only to the open vocabulary version of Detic
(\ie according to Detic terminology, no ``image label'' supervision, only ``caption'' supervision).

Furthermore, for a fair comparison as we use different backbones, detector architectures,
pretraining and self-training datasets, \etc,
we also reimplement the open vocabulary Detic, referred to as Detic$^\dagger$.
Our \emph{\threerec} approach and Detic$^\dagger$ are identical everywhere
apart from that pseudo-labelling is performed differently, where
\emph{\threerec} uses the most confident detection produced by \emph{\tworec} given the caption,
Detic$^\dagger$ uses the largest bounding box proposal from \emph{\tworec}.
It should be noted that our FCOS-based detector architecture is single-stage
and thus technically does not have bounding box proposals,
but we found that we can simulate producing proposals by filtering out the dense boxes
through thresholding the \emph{quality score};
we sweep a range of thresholds and pick the best performing one (0.75).
Our Detic$^\dagger$ reimplementation surpasses the original Detic
(\APall improved from 30.4\% to 34.8\%, and \APrare from 17.4\% to 23.4\%),
but our \emph{\threerec} approach beats this further (\APall of 35.7\% and \APrare of 25.6\%).

\tabLvisAll

\paragraph{Comparison with full LVIS training.}
In all other experiments detectors are trained on \LVISminusrare,
a version of the LVIS dataset where annotations for rare classes are removed
in order to test zero-shot detection.
Here we investigate the value of those annotations by training a
\emph{\tworec} detector on the entire LVIS (\ie without removing annotations for rare classes).
As expected,
\emph{\tworec} trained on full LVIS beats \emph{\tworec} trained on \LVISminusrare
when evaluated on LVIS itself, due its superior performance on the rare classes (Table~\ref{tab:lvisall}).
However, the effects of the dataset change are marginal
when transferring to COCO and Objects365,
and our self-trained \emph{\threerec} detector is still decidedly the best.
Therefore, pseudo-labelled data can be more valuable than ground truth annotations.
Furthermore, when using the NFNet-F0 backbone, \emph{\threerec} even beats the `cheating'
\emph{\tworec} on LVIS itself, but this result does not hold for the larger and better
NFNet-F6 backbone.

\section{Qualitative results}
\label{sec:app:qual}

\figQualThreeVsTwo
\figQualThreeParis
\figQualLvis
\figQualCoco
\figQualObj

\paragraph{Zero-shot detection.}
Figure~\ref{fig:qual:threevstwo} compares the zero-shot performance of
the \emph{\tworec} and \emph{\threerec} networks -- \emph{\tworec} is capable
of detecting some classes not seen during training, but \emph{\threerec}
is better.
Figure~\ref{fig:qual:threeparis} further showcases
\emph{\threerec}'s zero-shot capabilities.

\paragraph{Random samples from benchmark datasets.}
Figure~\ref{fig:qual:lvis} shows detections on a sample of LVIS validation
images.
Transfer to COCO and Objects365 is shown in Figure~\ref{fig:qual:coco}
and~\ref{fig:qual:o365}, respectively.
Kuo \etal~\cite{kuo2023fvlm} estimate the overlap between the training
set classes of \LVISminusrare and the classes of
COCO and Objects365 to be 91\% and 63\%, respectively.
Therefore, Figure~\ref{fig:qual:o365} also illustrates
impressive zero-shot capabilities of our method.

\paragraph{Failure cases.}
Instead of the typical failure cases such as
difficulty in detecting objects that are highly occluded, small, in unusual poses,
\etc,
here we focus on failure cases specific to zero-shot open vocabulary detection.
While exhibiting impressive zero-shot performance, we observe that
the network does not detect some common objects such as
houses, buildings, windows, bricks, trees, leaves, rocks, eyes, \etc.
We speculate that this is because images containing these objects are often
seen during training but never annotated with a bounding box,
thus the network learns they are not-an-object and does not assign
a high enough quality score.
There is also the philosophical question of what is an object,
for example, object parts such as eyes generally do not get detected
by our networks.
However, the system does detect wheels, as they are annotated in
the \LVISminusrare training set.
All of the above indicates that that there is room for improvement
in the zero-shot capabilities of the objectness component of detectors
(`quality score' here, region proposal networks in two-stage detectors),
or that this concept should be removed completely
as being somewhat incompatible with zero-shot open vocabulary detection.

\paragraph{Image attribution.}
Images used in \isArXiv{%
Figure~\ref{fig:teaser},~\ref{fig:qual:threevstwo}
}{%
Figure~\ref{fig:qual:threevstwo}}
and~\ref{fig:qual:threeparis}
were downloaded from Wikimedia%
\isArXiv{ or are personal photos of an author (Relja Arandjelović)},
and all are free to use and modify -- we made no modifications apart
from resolution change and the overlaying of the detections.
The Wikimedia originals can be found at the following links:

\begin{Verbatim}[fontsize=\tiny, breaklines=true, breakafter=\_, breakaftersymbolpre=none]
https://commons.wikimedia.org/wiki/File:Parisian_gargoyle_(Unsplash).jpg
https://commons.wikimedia.org/wiki/File:Venezia-gondola_on_canal_grande.JPG
https://commons.wikimedia.org/wiki/File:Macaroons_at_Smiths.jpg
https://commons.wikimedia.org/wiki/File:Round_hay_bales_and_a_hot_air_balloon_somewhere_in_Luxembourg.jpg
https://commons.wikimedia.org/wiki/File:POOL_HALL_-_NARA_-_543975.jpg
https://commons.wikimedia.org/wiki/File:Jumping_over_the_moon.jpg
https://commons.wikimedia.org/wiki/File:Wolves_chasing_a_wapiti,_Yellowstone_River_(2).jpg
https://commons.wikimedia.org/wiki/File:2020-01-11_Men's_Ice_hockey_3x3_Preliminary_round_Team_Blue_vs._Team_Orange_(2020_Winter_Youth_Olympics)_by_Sandro_Halank–057.jpg
\end{Verbatim}

\isArXiv{}{
{\small
\bibliographystyle{ieee_fullname}
\bibliography{bib/shortstrings,bib/more,bib/vgg_other,bib/vgg_local}
}

}{}

\end{document}